\documentclass{article}

\usepackage{microtype}
\usepackage{graphicx}
\usepackage{booktabs} %

\usepackage{hyperref}

\usepackage[accepted]{mlsys2022}

\mlsystitlerunning{Plumber: Diagnosing and Removing Performance Bottlenecks in Machine Learning Data Pipelines}

\usepackage{hyperref}
\usepackage{xfrac}
\usepackage{subcaption}
\usepackage{todonotes}
\usepackage{graphicx}
\usepackage{booktabs} %
\usepackage{xcolor}
\usepackage{wrapfig}
\usepackage{adjustbox}
\usepackage{tabularx}
\usepackage{makecell,rotating}
\usepackage{enumitem}
\usepackage{siunitx}
\usepackage{amsthm}
\usepackage{amsmath}
\usepackage{xspace}
\usepackage{soul}
\usepackage[utf8]{inputenc}
\usepackage[english]{babel}
\usepackage{blindtext}
\usepackage[htt]{hyphenat}
\usepackage{tikz} %
\DeclareFixedFont{\ttb}{T1}{txtt}{bx}{n}{12} %
\DeclareFixedFont{\ttm}{T1}{txtt}{m}{n}{12}  %

\usepackage{color}
\definecolor{deepblue}{rgb}{0,0,0.5}
\definecolor{deepred}{rgb}{0.6,0,0}
\definecolor{deepgreen}{rgb}{0,0.5,0}
\definecolor{codegreen}{rgb}{0,0.6,0}
\definecolor{codegray}{rgb}{0.5,0.5,0.5}
\definecolor{codepurple}{rgb}{0.58,0,0.82}
\definecolor{backcolour}{rgb}{0.95,0.95,0.92}
\definecolor{codeorange}{rgb}{1.0,0.64,0.0}

\usepackage{listings}

\newcommand\digitstyle{\color{codeorange}}
\makeatletter
\newcommand{\ProcessDigit}[1]
{%
  \ifnum\lst@mode=\lst@Pmode\relax%
   {\digitstyle #1}%
  \else
    #1%
  \fi
}

\newcommand\pythonstyle{\makeatother\lstset{
language=Python,
basicstyle=\footnotesize\ttm,
deletekeywords={map},
otherkeywords={self,@plumber},             %
emph={MyClass,__init__},          %
emphstyle=\color{deepred},    %
stringstyle=\color{deepgreen},
frame=tb,                         %
showstringspaces=false,            %
    commentstyle=\color{codegray},
    keywordstyle=\color{magenta},
    numberstyle=\tiny\color{codepurple},
    stringstyle=\color{codegreen},
    basicstyle=\ttfamily\footnotesize,
    breakatwhitespace=false,
    breaklines=true,
    captionpos=b,
    keepspaces=true,
    numbers=left,
    numbersep=5pt,
    showspaces=false,
    showtabs=false,
    tabsize=2,
    keywordstyle=[2]{\color{deepblue}},
    morekeywords=[2]{map, shuffle, batch, prefetch},
  literate=
    {0}{{{\ProcessDigit{0}}}}1
    {1}{{{\ProcessDigit{1}}}}1
    {2}{{{\ProcessDigit{2}}}}1
    {3}{{{\ProcessDigit{3}}}}1
    {4}{{{\ProcessDigit{4}}}}1
    {5}{{{\ProcessDigit{5}}}}1
    {6}{{{\ProcessDigit{6}}}}1
    {7}{{{\ProcessDigit{7}}}}1
    {8}{{{\ProcessDigit{8}}}}1
    {9}{{{\ProcessDigit{9}}}}1
    {<=}{{\(\leq\)}}1,
    morestring=[b]",
    morestring=[b]',
    morecomment=[l]//,
}}

\lstnewenvironment{python}[1][]
{
\pythonstyle
\lstset{#1}
}
{}

\newcommand\pythonexternal[2][]{{
\pythonstyle
\lstinputlisting[#1]{#2}}}

\newcommand\pythoninline[1]{{\pythonstyle\lstinline!#1!}} %

\usepackage{amsmath,amsfonts,bm}

\def\eqref#1{equation~\ref{#1}}

\def\1{\bm{1}}

\DeclareMathAlphabet{\mathsfit}{\encodingdefault}{\sfdefault}{m}{sl}
\SetMathAlphabet{\mathsfit}{bold}{\encodingdefault}{\sfdefault}{bx}{n}

\newcommand{\E}{\mathbb{E}}

 \graphicspath{{Figures/}}
\usepackage{xurl}

\newcommand\leftRightCrop[4]{\adjustbox{trim={#1\width} {0} {#4\width} {0},clip,center}{\includegraphics[width=\linewidth * \real{#3}]{#2}}}

\newcommand\TensorFlow{{\texttt{Tensor\-Flow}}}
\newcommand\Jax{{\texttt{JAX}}}
\newcommand\plumber{{\texttt{Plum\-ber}}}
\newcommand\autotune{{\texttt{AUTOTUNE}}}
\newcommand\heuristic{{\texttt{HEURISTIC}}}
\newcommand\tfdata{\texttt{tf.data}}

\newcommand\getnext{\texttt{Next}}  %
\newcommand\Dataset{\texttt{Dataset}}
\newcommand\Datasets{\texttt{Datasets}}
\newcommand\Iterator{\texttt{Iterator}}
\newcommand\Iterators{\texttt{Iterators}}

\newcommand\Google{Google}

\newcommand\MathCores{n_c}

\newcommand\MathSequentialSet{\mathbb{S}}

\newcommand\MathOperatorSet{\mathbb{D}}

\newcommand{\rateobserved}[1]{$R_{\text{obs}}$\xspace}
\newcommand{\rateprocessingraw}[1]{R_{\text{proc}}\xspace}
\newcommand{\rateprocessing}[1]{$R_{\text{proc}}$\xspace}
\newcommand{\rateprocessingrootraw}[1]{\hat{R}_{\text{proc}}\xspace}
\newcommand{\rateprocessingroot}[1]{$\hat{R}_{\text{proc}}$\xspace}
\newcommand{\rateestimated}[1]{$R_{\text{est}}$\xspace}
\newcommand{\rateestimatedraw}[1]{R_{\text{est}}\xspace}
\newcommand{\rateestimatedparallel}[1]{$R_{\text{estP}}$\xspace}
\newcommand{\rateestimatedparallelraw}[1]{R_{\text{estP}}\xspace}

\newcommand\leftExpResult[1]{\leftRightCrop{0.0155}{#1}{1.09}{0.01}}
\newcommand\rightExpResult[1]{\leftRightCrop{0.08}{#1}{1.09}{0.02}}

\newcounter{inlineenum}
\renewcommand{\theinlineenum}{\arabic{inlineenum}}
\newenvironment{inlineenum}
  {\unskip\ignorespaces\setcounter{inlineenum}{0}%
   \renewcommand{\item}{\refstepcounter{inlineenum}{\theinlineenum)~}}}
  {\ignorespacesafterend}

\newcounter{obscount}
\newcommand{\observation}{ %
\addtocounter{obscount}{1} %
\textbf{Observation \arabic{obscount}: }}

\newcommand*\circled[1]{\tikz[baseline=(char.base)]{
            \node[shape=circle,draw,inner sep=1pt] (char) {#1};}}

\begin{document}

\twocolumn[
\mlsystitle{Plumber: Diagnosing and Removing Performance Bottlenecks in Machine Learning Data Pipelines}

\mlsyssetsymbol{equal}{*}
\mlsyssetsymbol{oldinst}{*}

\begin{mlsysauthorlist}
\mlsysauthor{Michael Kuchnik}{cmu,oldinst}
\mlsysauthor{Ana Klimovic}{eth,oldinst}
\mlsysauthor{Jiří Šimša}{goo}
\mlsysauthor{Virginia Smith}{cmu}
\mlsysauthor{George Amvrosiadis}{cmu}
\end{mlsysauthorlist}

\mlsysaffiliation{cmu}{Carnegie Mellon University}
\mlsysaffiliation{eth}{ETH Zürich}
\mlsysaffiliation{goo}{Google}

\mlsyscorrespondingauthor{Michael Kuchnik}{mkuchnik@cmu.edu}

\mlsyskeywords{Machine Learning, MLSys}

\vskip 0.3in

\begin{abstract}
Input pipelines, which ingest and transform input data, are an essential part of training Machine Learning (ML) models.
However, it is challenging to implement efficient input pipelines, as it requires reasoning about parallelism, asynchrony, and variability in fine-grained profiling information.
Our analysis of over two million ML jobs in \Google{} datacenters reveals that a significant fraction of model training jobs could benefit from faster input data pipelines.
At the same time, our analysis indicates that most jobs do not saturate host hardware, pointing in the direction of software-based bottlenecks.
Motivated by these findings, we propose \plumber, a tool for finding bottlenecks in ML input pipelines.
\plumber{} uses an extensible and interpretable operational analysis
analytical model to
automatically tune parallelism, prefetching, and caching under host resource
constraints.
Across five representative ML pipelines, \plumber{} obtains speedups of up to
$47\times$ for misconfigured pipelines.
By automating caching, \plumber{} obtains end-to-end
speedups of over $50\%$ compared to state-of-the-art tuners.
 \end{abstract}
]

\printAffiliationsAndNotice{\mlsysWasAtGoogle} %

\section{Introduction}%
\label{sec:introduction}%
The past decade has witnessed tremendous advances in Machine Learning (ML), leading to custom hardware accelerators~\cite{jouppi2020domain}, sophisticated distributed software~\cite{tensorflow2015-whitepaper}, and increasing dataset sizes~\cite{krizhevsky2012imagenet,wu2016google,deng2009imagenet,OpenImages2,lin2014microsoft}.
While ML accelerators can radically reduce the time required to execute training (and inference) computations, achieving peak end-to-end training performance also requires an efficient input pipeline that delivers data for the next training step before the current step completes.
For example, an ImageNet dataloader can improve end-to-end ResNet-50 training
time by up to 10$\times$ by properly leveraging parallelism, software pipelining, and static optimizations~\cite{murray2021tfdata,mattson2019mlperf,he2016deep}.
An efficient input pipeline also ensures that accelerator hardware is well-utilized, lowering costs.

Our analysis of over two million ML training jobs from a variety of domains (e.g., image recognition, natural language processing, reinforcement learning) at \Google{} shows that input data processing bottlenecks occur frequently in practice, wasting valuable resources as ML accelerators sit idly  waiting for data.
We find that in 62\% of jobs, the input data pipeline repeatedly produces
batches of data with a delay of at least 1ms \emph{after} the accelerator/model
is able to consume it, incurring a non-negligible slowdown per training step.

To understand this phenomenon, we classify input pipeline bottlenecks as \emph{hardware bottlenecks}, which occur when input data processing saturates host CPU and/or memory resources,
or \emph{software bottlenecks},
which don't saturate the host due to poor configuration or I/O.
We find that the majority of input data stalls arise due to software
bottlenecks, indicating a mismatch between host resources and the software that
drives them.
While today's input data pipeline libraries hide implementation complexity
behind easy-to-use APIs (\S~\ref{subsec:background-architecture}), it is
difficult for ML users to understand and optimize the performance properties of
input data pipelines.
Localizing an input pipeline bottleneck with existing systems requires profiling
the ML job and manually analyzing its execution trace
(\S~\ref{subsec:background-bottlenecks}), which is error-prone, burdensome, and
difficult to understand.
Our experience echoes lessons learned in databases and analytics, which indicate it is unreasonable to expect experts in ML to \textit{also} be experts in writing data pipelines~\cite{chamberlin1981history,olston2008pig,armbrust2015spark,fetterlydryadlinq,melnik2010dremel,olston2008automatic}.

Motivated by the prevalence of input pipeline bottlenecks across real ML
training jobs and the burden users experience when attempting to debug the
performance of input pipelines, we introduce \plumber, an open-source tool that traces
input pipeline execution, models the pipeline's components, and predicts the
effect of modifications.
\plumber{} can be used with a single line of code over arbitrary input
pipelines and consists of two components: a \textit{tracer} and a
\textit{graph-rewriter}.
Similar to database execution
planners~\cite{ioannidis1996query,oracle_explain_plan},
\plumber{}'s tracer quantifies the performance of individual operators, focusing the
practitioner's attention on
the most underperforming subset of the data pipeline, while also quantifying
the resource utilization (i.e., CPU, disk, memory) of the pipeline.
\plumber{}'s rewriter is an automatic front-end to the tracer and acts as an
optimizer without user intervention---introducing parallelism, caching, and prefetching in a
principled fashion, and can be extended to support more.
Our contributions are:%
\vspace{3pt}
\begin{enumerate}[nosep,wide=0pt,label={\textbf{(\arabic*)}}]%
\setlength{\itemsep}{3pt}
\item We analyze \textit{two million} ML jobs, providing evidence that input data
    processing is a common bottleneck (\S\ref{sec:analysis}).

  \item We introduce a principled tracing methodology, \textit{resource
    accounted rates} (\S\ref{sec:resource_accounted_rates}), which automatically estimates
    pipeline CPU, disk, and memory requirements.

  \item We present a novel linear programming (LP) (\S\ref{sec:linear_programming}) formulation using the rates traced
    during runtime, predicting an upper bound on performance.
    Unlike state-of-the-art tuners, which have unbounded error, the LP's predictions of system state
    are bounded within $4\times$
    by resource usage.

  \item We present \plumber{} (\S\ref{sec:system}), a tool that detects and
    removes input pipeline bottlenecks using resource-accounted rates and the LP\@.
    \plumber{} currently supports automatic injection of
    parallelism, prefetching, and caching, and paves a way forward toward more
    general query optimizer extensions. \plumber{} requires one line of code to
    use.

  \item We evaluate \plumber{} (\S\ref{sec:evaluation}) on five
    workloads with end-to-end performance improvements up to $47\times$ over
    misconfigured pipelines and 50\% over state-of-the-art tuners.
\end{enumerate}
\section{Plumbing Basics}%
\label{sec:background}%
All ML training begins with input data, which is curated by
input pipeline frameworks.
In this section, we outline the abstractions provided by input pipeline
frameworks, noting design decisions that have an effect on understanding
performance (\S\ref{subsec:background-architecture}).
We next jump into common tools for understanding bottlenecks
(\S\ref{subsec:background-bottlenecks}).

\subsection{Input Pipeline Architecture}%
\label{subsec:background-architecture}
Input pipelines specify: a data source, transformation functions, iteration orders, and grouping strategies.
For example, image classification pipelines read, decode, shuffle, and batch
their $($\texttt{image}, \texttt{label}$)$ tuples, called \textit{training examples}, into fixed-size arrays~\cite{deng2009imagenet,krizhevsky2012imagenet}.
Unlike batch processing frameworks (e.g., Spark~\cite{armbrust2015spark},
Beam~\cite{beam}, Flume~\cite{flume}), which may be used to \textit{create} the
data used for training, the main goal of input pipeline frameworks is to
dynamically \textit{alter} the training set online.
Three major reasons for online processing are:
\begin{inlineenum}
  \item data is stored compressed in order to conserve storage space,
  \item data is randomly altered online using data augmentations, and
  \item practitioners may experiment with features throughout modeling.
\end{inlineenum}

Input pipelines are programmed \textit{imperatively} or
\textit{declaratively}, each with different APIs.
Imperative frameworks, like PyTorch's and
MxNet's~\cite{pytorch-dataloader,mxnet_dataio} \texttt{DataLoader},
allow users to specify their pipelines in plain
\texttt{Python} by overriding the \texttt{DataLoader}.
Declarative libraries, like DALI~\cite{nvidia-dali} and \TensorFlow's
\texttt{tf.data}~\cite{murray2021tfdata,tfdata}, compose
functional, library-implemented primitives, which are executed by the
library's runtime.
While both styles are equally expressive in terms of pipeline construction,
frameworks leave a large part of implementation to the user.
In contrast, the libraries decouple specification from implementation,
requiring
the user to merely declare the pipeline structure, offloading optimizations
to the runtime.
We focus our discussion on \tfdata{}, because it allows for various backends to
service similar pipeline definitions, enabling tracing and tuning \textit{behind the
API}, and can be used with all major training ML frameworks.

\begin{figure}
  \pythonexternal{Python/example.py}
  \vspace{-10pt}
  \caption{Python pseudo-code for ImageNet-style training.
  Line 2 is file reading, line 3 is user-defined image processing, and line 4 samples, batches, and prefetches data. Lines 5--6 are the critical path of training, instantiating an \Iterator.
  }%
  \label{fig:pseudo_code}%
  \vspace{-10pt}%
\end{figure}

\begin{figure}
\centering%
\includegraphics[width=0.95\linewidth]%
{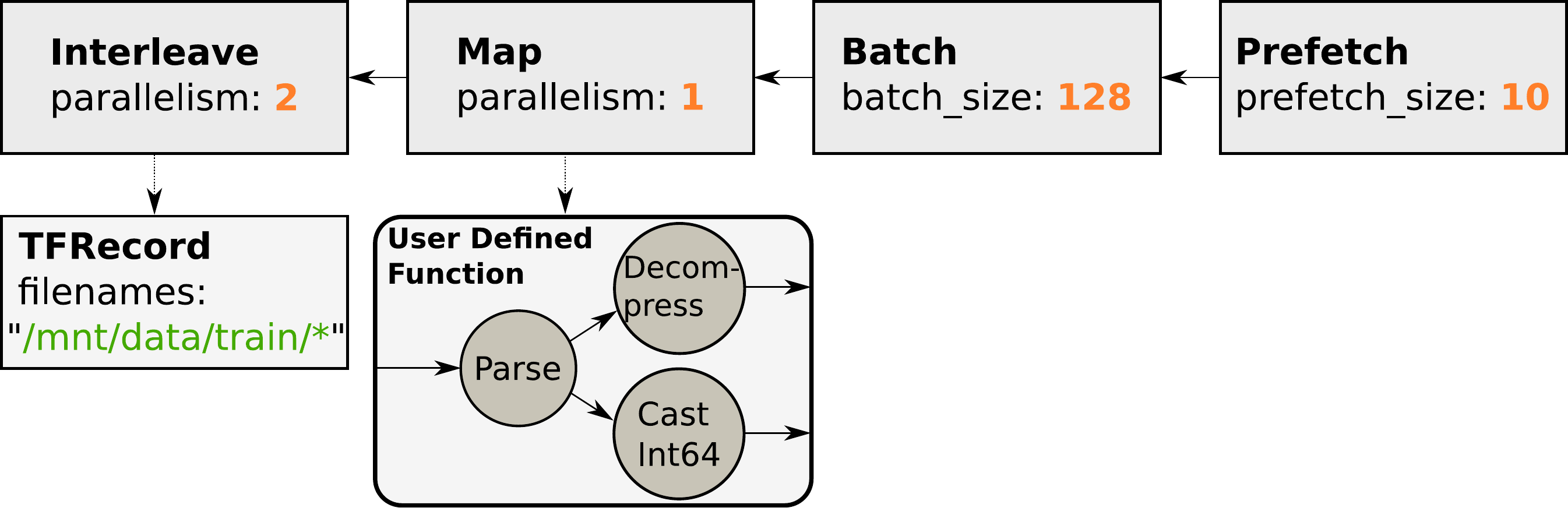}
\includegraphics[width=0.95\linewidth]%
{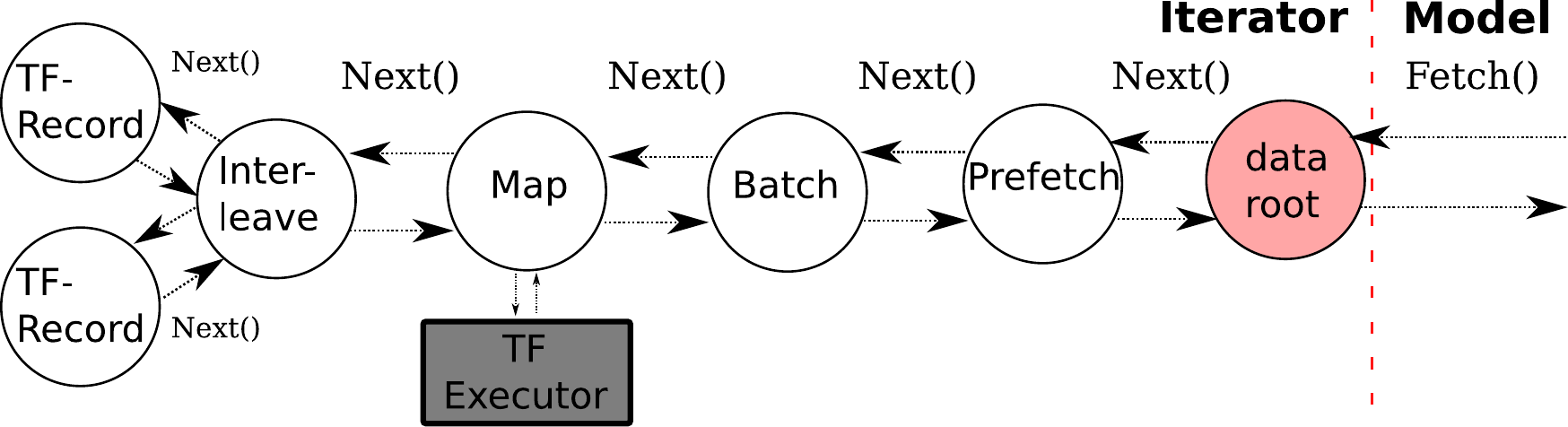}%
\vspace{-0.3cm}
  \caption{\textbf{Top (\Dataset{} View):} A \tfdata{} pipeline is composed of
  \Dataset{} objects characterized by attributes, e.g., by their level of parallelism.
  \textbf{Bottom (\Iterator{} View):} The root \Dataset{} is instantiated into
  an \texttt{Iterator} tree at runtime, which feeds the model.
  \Iterators{} pull data from their children in a recursive manner.
}%
\label{fig:tfdata_graph}%
\vspace{-15pt}
\end{figure}

\textbf{Input Pipeline Abstractions.}
In \tfdata{}, \Datasets{} are the basic building blocks for declaring input pipelines.
In Figure~\ref{fig:pseudo_code}, each function call chains \Datasets{}.
Instantiating a \Dataset{} (line 5) yields a tree composed of one or more \Iterators{}, which produces a sequence of training examples through an iterator interface that maintains the current position within a \Dataset{}.
Figure~\ref{fig:tfdata_graph} illustrates how \Datasets{} are unrolled into an \Iterator{} tree. Some \Datasets{} (see \texttt{Map}) implement multi-threaded parallelism within the corresponding \Iterator, while others (see \texttt{TFRecord}) can only be parallelized by reading from multiple sources in parallel (e.g., using \texttt{Interleave}).
An \texttt{Iterator} implements the following three standard Iterator model~\cite{XRM,volcanodb} methods:
\begin{itemize}[noitemsep,topsep=2pt,leftmargin=*]
    \item \textbf{Open} defines \Iterator{} parameters and references to child \Iterators{} and initializes internal state.

    \item \textbf{Next} yields an example from the \Iterator{} or a signal to
      end the stream. Source nodes read from storage or memory to yield
      examples. Internal nodes call \getnext{} on their children to gather examples before applying transformations to them. In Figure~\ref{fig:pseudo_code}, the data source outputs file contents, which then undergo image processing, shuffling, and batching.

    \item \textbf{Close} releases resources and terminates the \Iterator{}.
\end{itemize}

\textbf{User-Defined Functions.}
User-defined functions (UDFs)
comprise the bulk of data pipeline execution time and are used to implement
custom data transformations.
Figure~\ref{fig:pseudo_code} shows an example of a computer-vision
pipeline utilizing UDFs, which perform image
decoding, image preprocessing, and tensor transposing for efficient execution on
accelerators.
Users are able to write UDFs in a restricted form of \texttt{Python},
which is compiled into an efficient and parallel implementation.
\subsection{Understanding Input Bottlenecks}%
\label{subsec:background-bottlenecks}
An input bottleneck occurs when the input pipeline is not able to generate batches of training examples as fast as the training computation can consume them.
If the time spent waiting for the input pipeline
exceeds tens of microseconds on average, the input pipeline is not
keeping up with model training, causing a \textit{data stall}~\cite{dsanalyzer}
The current practice of pipeline tuning, which optimizes the throughput (rate)
of the pipeline, is explained below.

\textbf{Profilers.} 
Event-based profilers, such as the
\TensorFlow{} Profiler~\cite{TFProfiler}, can emit metadata at
particular software events to aid in determining control-flow.
As there are \textit{thousands of concurrent events per second} for a pipeline,
it is difficult to quantitatively determine
which events actually \textit{caused} a throughput slowdown.
To automate and generalize past heuristics deployed in guides~\cite{tfdata_xprof_playbook},
the \TensorFlow{} Profiler added a bottleneck discovery feature~\cite{TFDataBottleneckAnalysis}.
This tool works by finding the \Iterator{} with highest impact on the
\textit{critical path} of a \Dataset{}.
However, it can only rank \Datasets{} by slowness,
and it can't predict their effect on performance.
Furthermore, critical-paths are not well-defined for concurrent and randomized
event graphs (e.g., execution times of individual operations overlap and are
data-dependent), forcing heuristics to be used.

\textbf{Tuners.}
\tfdata{} applies dynamic optimization of pipeline parameters when users specify
\autotune{} for supported parameters,
such as the degree of parallelism and size of prefetch buffers~\cite{murray2021tfdata}.
The autotuning algorithm works by representing \Iterators{} in a pipeline as an
M/M/1/k queue~\cite{queue_book,lazy_repairman},
and a formulation for the queue's
latency is analytically determined.
Statistics about the execution of \Iterators{} are
recorded with a lightweight harness.
Combining the analytical model with the runtime statistics enables tuning the latency of the pipeline with
respect to performance parameters.
Specifically, the processing time of each element is normalized by the
parallelism and the ratio of input to output elements.
This statistic is then combined with ``input latency''
statistics of the children nodes in a
node-type dependent way to get an ``output latency''.
Output latency tuning is done via hill-climbing or gradient descent and
ends when the tuning plateaus or reaches a resource budget.
While \autotune{} works in practice,
it is hard to understand and extend for two reasons.
First, open-systems, like M/M/1/k queues, have a throughput purely dependent on
input arrival rates, which are not applicable for closed-systems.
Second, because resource utilization is not modeled,
the output latency function can be driven to zero if parallelism is allowed to
increase unbounded, forcing heuristic constraints to be used.
\section{Spot the Leak: Fleet Analysis}%
\label{sec:analysis}%
We analyzed over two million ML jobs that ran in \Google{} datacenters to
determine whether input bottlenecks are common and characterized their most typical root cause.
The jobs we analyzed used \tfdata{} and ran over a one-month period from July to August of 2020.
The workload included production and research jobs from a variety of ML domains, including image recognition, natural language processing, and reinforcement learning.
To measure the frequency of input bottlenecks in practice, we measured the average time spent fetching input data per training step across jobs.

\subsection{Are Input Bottlenecks Common?}
We detect input pipeline bottlenecks by measuring the average latency across all
\Iterator{} \getnext{} calls, which is the average time the job spends blocked waiting for input data in each training step.

\observation{\textit{For a significant fraction of ML jobs, the input data pipeline produces data at a slower rate than the model is able to consume it.}}  

\begin{figure}%
\centering%
\includegraphics[width=0.99\linewidth]{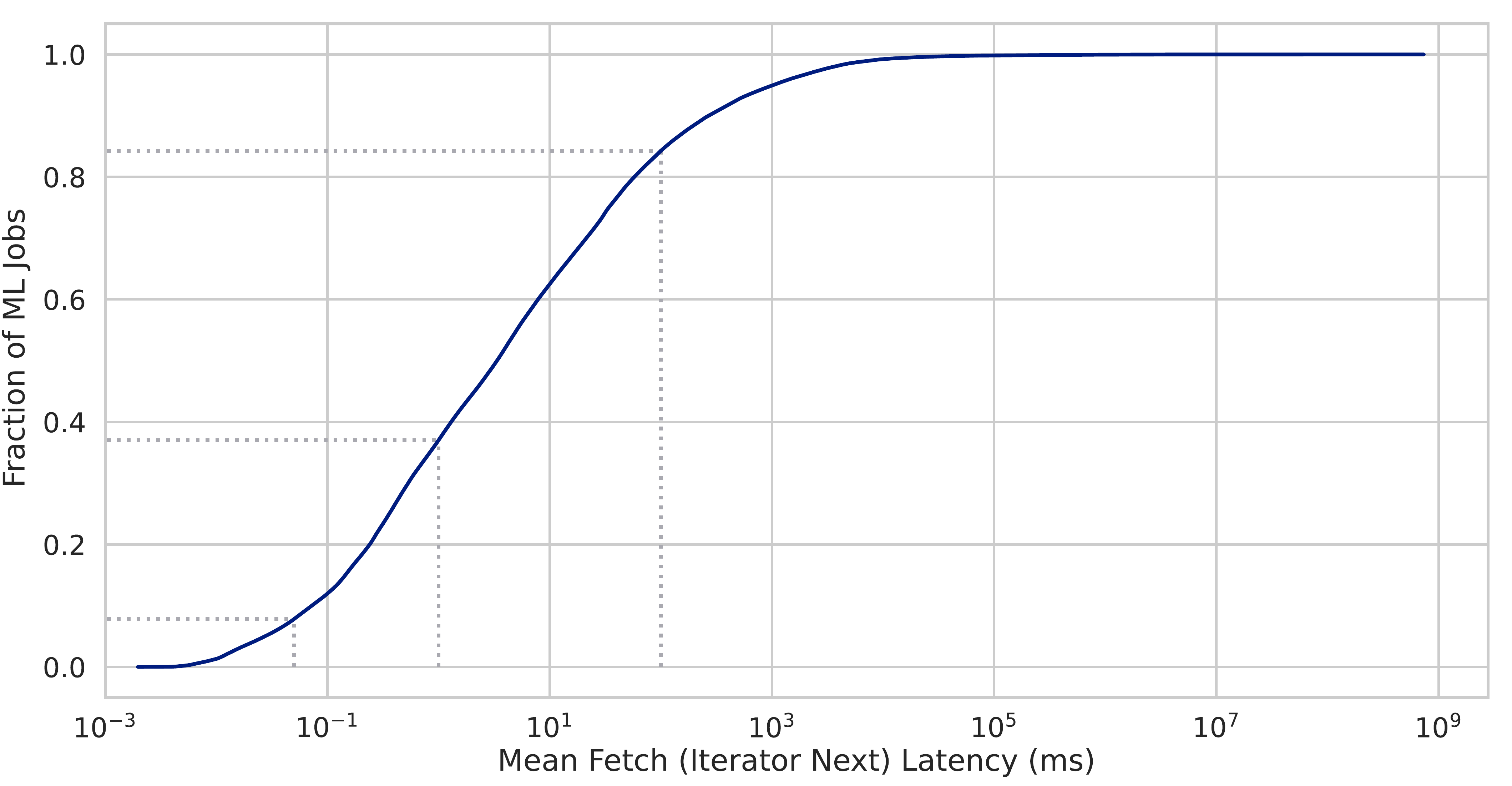}%
\vspace{-0.3cm}
  \caption{Time spent fetching training examples using \getnext{}, in ms. On
  average, for 92\% of jobs \getnext{} latency exceeds 50$\mu$s, for 62\% of
  jobs it exceeds 1ms, and for 16\% of jobs it exceeds 100ms. Fetch latencies for well-configured pipelines are in the low tens of \textit{microseconds}.
}%
\label{fig:mean_duration_ms_hist}%
\vspace{-10pt}
\end{figure}

Figure~\ref{fig:mean_duration_ms_hist} shows that for 62\% of jobs, the average
\getnext{} latency per training step exceeds 1ms and for 16\% of jobs, the average wait time exceeds 100ms.
Since the input pipeline affects both end-to-end training performance and
hardware accelerator utilization, it is a critical part of ML training to optimize.
We note that fetch latency applies to \textit{each iteration} of the
training process---over ten thousand times in a typical session.
At any point in time, between 1--10\% of the fleet is waiting on input data,
which is significantly costly at the scale the fleet operates at.

\subsection{Why Do Input Bottlenecks Occur?}
We classify input pipeline bottlenecks into two categories: hardware and software bottlenecks. \textit{Hardware bottlenecks} occur when hardware resources used for data processing saturate.
The hardware resources typically used for input processing are CPU cores, host memory, and local or remote storage.
Many workloads do not use local storage for I/O, but pull their data from
external data sources (e.g., distributed filesystems)~\cite{murray2021tfdata}.
Thus, external I/O resources may bottleneck training jobs.
Hardware resource saturation can be remedied by adjusting the resource allocation, e.g., switching to a node with more cores, memory, or I/O bandwidth.

\textit{Software bottlenecks} occur because the software is not driving the hardware efficiently, e.g., by using too little or too much parallelism, or incorrectly sizing prefetch buffers to overlap communication and computation.
When a user is confronted with a software bottleneck, they must find the root
cause and fix it; otherwise, they risk underutilizing hardware performance.
We note I/O bottlenecks can also be caused by software configuration, due to
inefficient access patterns and low read parallelism.

\observation{\textit{Host hardware is rarely fully utilized for jobs with high input pipeline latencies. Thus, input bottlenecks are likely rooted in software or I/O inefficiencies, rather than hardware saturation.}}

\begin{figure}%
\centering%
\includegraphics[width=0.99\linewidth]{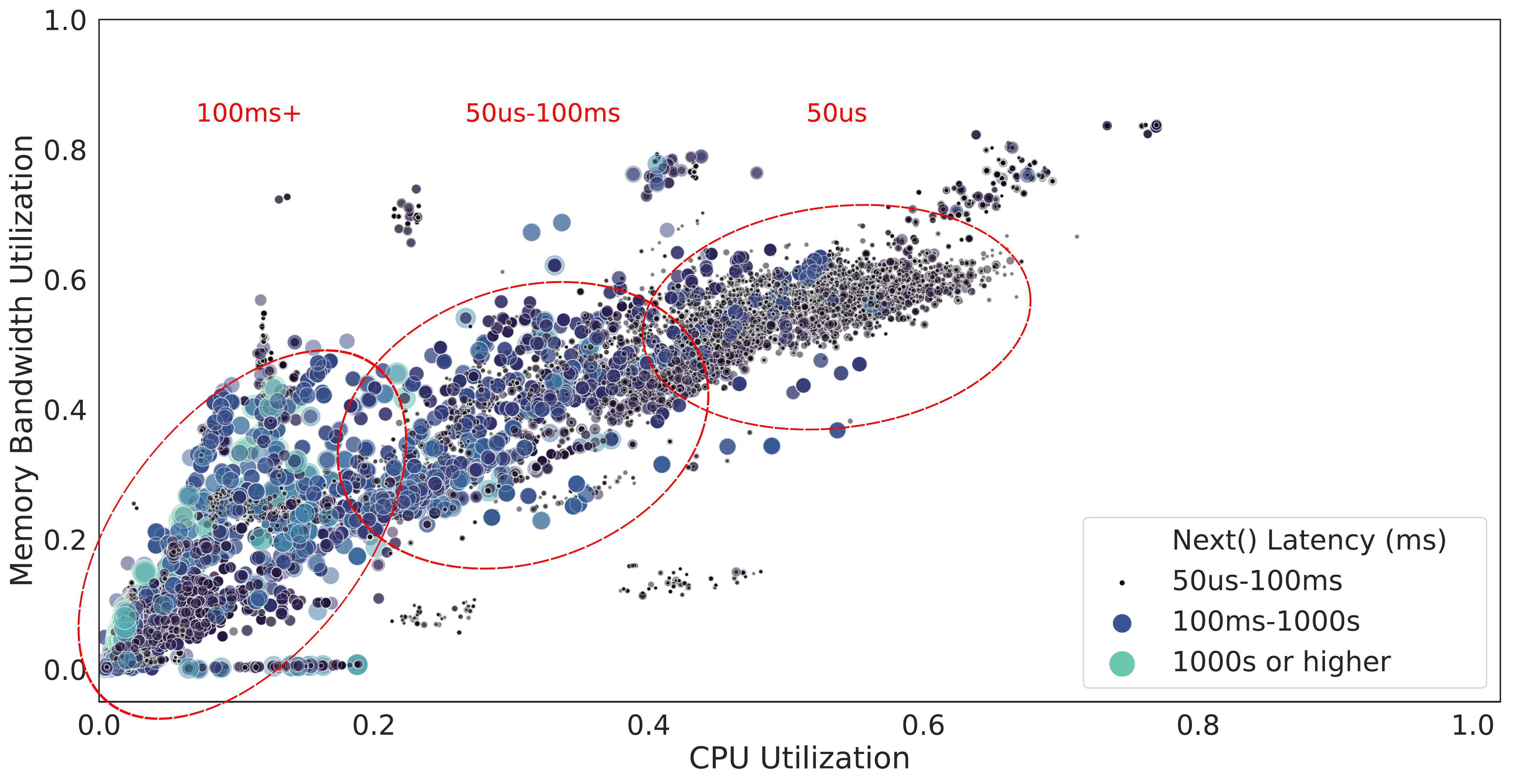}%
\vspace{-0.3cm}
\caption{CPU utilization of training jobs compared to their memory bandwidth
  utilization. Larger points are for jobs with longer pipeline latency.
  We annotate three major clusters.
  The
  average CPU and memory-bandwidth usage is 11\% and 18\%, respectively,
  for jobs with pipeline latency of 100ms or more.
  The majority of jobs do not saturate host resources, suggesting bottlenecks in software.
}%
\label{fig:resource_usage}%
\vspace{-10pt}
\end{figure}

To understand the breakdown between these categories of input bottlenecks,
we measure the host CPU and memory bandwidth resource utilization for the jobs captured in our analysis.
In Figure~\ref{fig:resource_usage}, we show a breakdown of different jobs'
average \getnext{} call latencies organized according to the CPU and memory bandwidth utilization of the pipeline host.
We exclude jobs with latency below $50\mu$s because for those jobs the input pipeline is not a bottleneck, as it takes tens of microseconds to read input data that is readily available from a prefetch buffer (including thread wakeup and function invocation time).
Our data indicates that jobs with latency between $50\mu$s and 100ms (small, dark dots) utilize \emph{more} of the host's resources than those with latency higher than 100ms (large, blue dots).
For context, a TPUv3-8~\cite{jouppi2020domain,MLPerf07}
takes roughly 120ms to process a minibatch for ResNet-50~\cite{he2016deep}.
Jobs where the average fetch latency exceeds 100ms are, then, \textit{significantly} input-bound, and as shown in the Figure~\ref{fig:resource_usage} their resource usage for both memory bandwidth and CPU is concentrated below 20\%.
\section{Plumber}%
\label{sec:system}%
To address the challenges that users face in configuring input data pipelines in ML jobs, we introduce \plumber{},
an extensible tool that automatically detects and removes input data bottlenecks.
Using a top-down approach, we start with the architecture of \plumber{}
(\S\ref{subsec:architecture}) and then explain tuning methodology
(\S\ref{subsec:tuning}).
We later formulate the resource allocation problem across \Datasets{} as a
LP (\S\ref{sec:linear_programming}), which relies on
per-\Dataset{} resource rates derived by \plumber{}
(\S\ref{sec:resource_accounted_rates}).
\plumber{} is released as an open-source artifact (\S\ref{sec:artifact}) and consists of
a 3k line patch on top of \tfdata{}'s C++ \autotune{} infrastructure and 8k
lines of \texttt{Python} interface.

\subsection{Software Architecture}%
\begin{figure}%
\centering%
\includegraphics[width=0.99\linewidth]{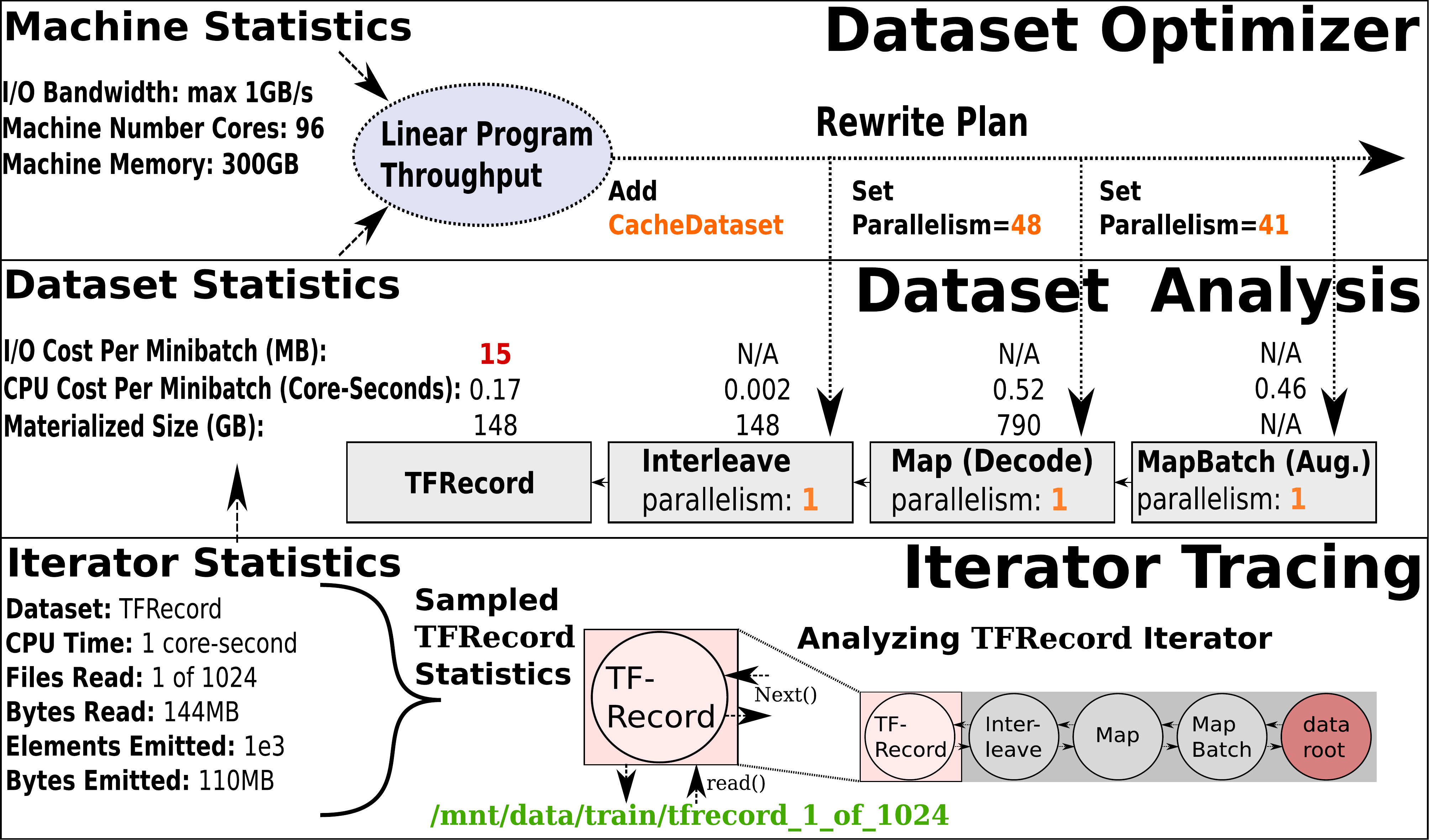}%
  \vspace{-0.5em}
  \caption{%
  An ImageNet pipeline with Plumber's various states of processing.
  \plumber{} starts with \Dataset{}-level tracing, which is then followed by
  analysis for CPU, disk, and memory costs, which are subsequently modeled and
  optimized.
  Reading a \texttt{TFRecord}, for example, is converted from bytes read to an
  I/O cost per minibatch, which can then be linked with available I/O resources
  to find bottlenecks.
  \vspace{-0.5em}
  }%
  \label{fig:plumber_example_overview}%
  \vspace{-10pt}
\end{figure}

\label{subsec:architecture}%
\plumber{} reasons about performance in a \textit{layered} fashion.
The goal of the layers is to abstract basic \Dataset{}-level statistics into
costs, which can be compared, optimized over, and extended.
We demonstrate this architecture with a simplified, misconfigured
ImageNet pipeline in Figure~\ref{fig:plumber_example_overview}.
This pipeline requires reading from \texttt{TFRecords} in
parallel, decoding the examples of each record, and randomly augmenting
the examples with crops and flips.
Each of the pipeline components has a tunable, except for the
\texttt{TFRecords}, which is parallelized by \texttt{Interleave}.

\textbf{Tracing.}
By enabling a runtime flag, tracing collects \Dataset{}-level statistics, such as counters for
elements processed, CPU time spent, and the number of bytes per element.
\plumber{} periodically dumps these statistics into a file along with
the entire serialized pipeline program.
Joining the \Datasets{} with their program counterpart enables building an
in-memory model of the pipeline dataflow.
The bottom of Figure~\ref{fig:plumber_example_overview} shows how a
\texttt{TFRecord} \Dataset{} is traced, which reads a single \texttt{TFRecord} file.
\plumber{}'s tracing instruments all \texttt{read()} calls into the filesystem
within \tfdata{},
allowing it to see all the reads into the 144MB file.
Each record is unpacked into roughly 1200 elements (110kB images), which \plumber{} counts.
By inspecting the serialized program, \plumber{} knows that there
are 1024 total \texttt{TFRecord} files, and thus can estimate that the dataset
contains $1024 \times 1200$ elements---ImageNet has 1.2 million.
To get CPU usage, \plumber{} wraps a thread-level CPU timer around
\getnext{} calls, which only counts \textit{active} (not blocked)
CPU-cycles.
All \getnext{} calls are instrumented such that:
\begin{inlineenum}
  \item CPU timers stop when \Datasets{} call into their children and start when control is returned and
  \item statistics (e.g., counts and sizes) about each yielded element are attributed to the producer.
\end{inlineenum}
The statistics necessary for optimization total less than 144 bytes per
\Dataset{}.

\textbf{Analysis.}
To find bottlenecks, \plumber{} must analyze the
traced data using analytical modeling, which puts each \Dataset{} in units of
cost that can be compared.
\plumber{} treats the pipeline as a closed system, with each component operating
asynchronously yet sharing the same resource budget.
For example, we can see that \texttt{TFRecord} reads at a rate of 15MB per
minibatch, but
decoding consumes 1/2 core per minibatch.
Determining which \Dataset{} is the bottleneck depends on the resource allocation of CPU and
I/O---for example, 30 minibatches per second can only be hit with
450MB/s of I/O and 15 CPU cores.

\textbf{Optimizer.}
\plumber{} supports reasoning about CPU and disk
parallelism, caching, and prefetch injection.
Achieving optimal CPU and I/O parallelism requires allocating sufficient parallelism to keep the
pipeline balanced in terms of throughput capacity, which we formalize in the
LP\@.
Caching reduces the amount of work done by avoiding re-computing data
dependencies, and the optimal cache minimizes total work by placing it as high
in the pipeline as possible.
Prefetching is a subsequent pass which injects prefetching
proportional to the idleness in the pipeline under a benchmark workload.
As shown in the example, by placing caching at \texttt{MapDecode},
we can avoid all CPU and I/O operations except
for the subsequent Crop and Batch at the expected cost of 790GB of memory.
In this example, the optimizer knows that the machine only has 300GB of memory
and thus it must settle with caching at the 148GB \texttt{Interleave}.
By caching \texttt{Interleave}, \plumber{} can
redistribute remaining CPU resources to other stages with the LP\@.

\textbf{Extensions.}
Extending \plumber{} requires minimal effort by cleanly interacting with existing
optimizations.
For example, to add the ability to cache materialized results to disk in
addition to memory, one can reuse all caching logic up to the cache decision
itself, which would dispatch to in-memory caching preferably and disk caching
if space and disk bandwidth allow it.
Significant extensions (e.g., networking or GPU data
transfer I/O) require adding the corresponding tracing and rates
at the lower levels, which can then be incorporated as LP expressions
and constraints.
Two areas of future work are in extending \plumber's optimizer to
reason about correctness optimizations, such as reusing
data~\cite{data-echoing}, and extending \plumber{} to perform optimal resource
provisioning for matching a target throughput (e.g., to minimize cost).
 
\subsection{Tuning Methodology}%
\label{subsec:tuning}
The intuition behind \plumber{}'s performance debugging methodology is that input pipeline performance is limited by the \Dataset{} with the lowest throughput and can be alleviated by adjusting the resource allocation (out of the available hardware resources) for this \Dataset.
However, simply maximizing the number of threads used to compute elements for a parallelizable \Dataset{} is not always productive, as threads compete for CPU and memory resources. 
The end-to-end throughput of the input pipeline can also be limited by I/O
bandwidth. Caching \Datasets{} in memory alleviates CPU or I/O bottlenecks at the cost of memory capacity. 
As all \plumber{} traces are also valid programs (that can be rewritten),
\plumber{} simply requires a way for a user to mark the program for tracing, and
thus one entry point is sufficient for \plumber{} to accomplish all tuning.

\textbf{Tuning Interface.}
\plumber{} introduces \textit{tuning annotations} to add on top of existing
data loading code.
A data loading function is one which returns a \tfdata{} \Dataset, which has an
associated signature.
Tagging the code with an \texttt{@optimize} annotation gives \plumber{}
permission to 
intercept one or more \Datasets{} and return an optimized variant matching the 
\Dataset{} signature.
The annotation provides \plumber{} with an entry point into the loader allowing
\plumber{} to trace it under a benchmark workload, and
rewrite it before passing it back to the application.

\textbf{Modeling.}
\plumber{} maximizes throughput by
modeling the input pipeline as an asynchronous closed-system in the context of operational
analysis~\cite{denning1978operational}, which explicitly defines bottlenecks.
The operational framework has few statistical assumptions, unlike Markovian queues, and parameterizes each component
of a system with a cost relative to the resource usage, usually expressed in
units of time.
\plumber{} further measures the resources and traces the
analytical network to automatically ``operationalize'' and tune an arbitrary input
pipeline.

\subsection{Allocating Hardware Resources}%
\label{sec:linear_programming}%
In order to derive an improved input pipeline configuration,
we need to understand how each \Dataset's performance is affected by changing the fraction of hardware resources allocated to it, e.g., by changing the degree of parallelism of a \Dataset{} or inserting prefetching or caching \Datasets{} which consume extra memory.
We are interested in characterizing the usage of three types of resources in input pipeline execution: CPU, disk bandwidth, and memory capacity.
We first formulate a Linear Program (LP) to solve for the optimal CPU resource
allocation before moving onto disk and, finally, memory capacity.
The calculation of the inputs into this LP are discussed in the following
subsection (\S\ref{sec:resource_accounted_rates}).

\textbf{CPU.}
For CPU optimizations, we optimize over a \Dataset{} \textit{tree} in the input
pipeline to decide \textit{what fraction} of the CPU-time each \Dataset{} should
receive such that throughput is maximized.
Having two \Datasets{} in series with rate $R_1$ and $R_2$ yields an aggregate rate of $X=\text{min}(R_1, R_2)$. The bottleneck \Dataset{} determines performance.
For example, in Figure~\ref{fig:tfdata_graph}, if \texttt{Map} has rate $R_1$
and \texttt{Batch} has rate $R_2$, to increase $X$, we must assign the slower of the two \Datasets{} more resources.
However, the above is only true \textit{if} we can parallelize the bottleneck node and \textit{if} we have resources left.

We optimize $\text{Max}_\theta \left[ {X=\min_{i\in \MathOperatorSet}[\theta_i * R_i]} \right]$ subject to constraints:
$\sum_{i \in \MathOperatorSet}{\theta_i} \leq \MathCores$;
$\theta_{i \in \MathOperatorSet} \geq 0$;
$\theta_{i \in \MathSequentialSet} \leq 1$.
In the above equations, $\MathOperatorSet$ is the set of \Datasets{} in consideration,
$\MathSequentialSet \subseteq \MathOperatorSet$ a
subset of sequential \Datasets, $\MathCores$ the total number of cores, $X$ throughput, $R_i$ the measured rate of minibatches per second per core, and $\theta$ the fractional number of cores allocated.
The equation maximizes the input pipeline throughput $X$ by maximizing the aggregate rate ($\theta_i * R_i$) of the slowest \Dataset{} (the minimum).
We constrain sequential \Datasets{} to have at most one core, and we cannot exceed all cores on the machine.

\textbf{Disk.}
As shown in Figure~\ref{fig:tfdata_graph}, data flows from disk (\texttt{TFRecordDataset}) into the CPU section of the pipeline.
Therefore, if the data source is slower than the rest of the pipeline, the entire pipeline is
disk-bound.
For large reads (e.g., records), there are two factors that can cause a disk
bound:
\begin{inlineenum}
  \item insufficient read parallelism and
  \item max disk I/O bandwidth.
\end{inlineenum}
The latter can be found via benchmarking tools~\cite{fio}, but \plumber{} goes a
step further by benchmarking entire empirical parallelism vs.\ bandwidth curve
for a data source (via rewriting).
The source parallelism results can then be fit with a piecewise linear curve to
be injected into the optimizer to determine a minimal parallelism to hit max
bandwidth.

\textbf{Memory.}
Caching aggressively is always desirable, because it allows I/O
to be entirely reduced and some CPU processing to be partially reduced.
The potential speedup grows as one goes up the pipeline, but so does the memory
requirement to cache, which may exceed system memory.
\plumber{} checks if operators are random
functions (e.g., augmentations), as described in the appendix
(\S\ref{subsec:random_udfs})---randomized functions have infinite cardinality
and cannot be cached.
However, if an operator's output is finite, \plumber{} estimates the
materialized size.
To solve memory caching for typical, linear-structure pipelines,
\plumber{} uses a greedy (yet optimal) approach to select the \Dataset{} closest to the root that fits in memory.
\plumber{} can also solve for more generic topologies by adding Boolean decision variables for each cache candidate over the LP already presented.

\subsection{Resource Accounted Rates}%
\label{sec:resource_accounted_rates}
The LP construction (\S\ref{sec:linear_programming}) assumes the existence of calculated
rates (for CPU and disk I/O) as well as the materialized size (for caches), and
uses those values as inputs into the algorithm.
This section explains how those values are calculated by using resource
accounted rates.
Resource accounted rates encompass the cost of an
operation or decision to cache in the pipeline.
For CPU and I/O, the cost is the ratio of CPU core-time or I/O bytes per minibatch,
which are \textit{throughput bounds}.
Memory capacity measures the cost in terms of bytes required for
materialization at a particular \Dataset{}, and is a \textit{throughput
optimization}.
The middle of Figure~\ref{fig:plumber_example_overview} illustrates all three
of these costs.
The full algorithm for resource accounted rates can be found in the appendix;
however, we give a brief description here.

\textbf{Common Units.}
The root of the pipeline gives a common set of units for CPU and I/O\@:
minibatches.
Children of the root do not necessarily output elements in terms of minibatches
(e.g., prior to batching); thus, a conversion factor between an arbitrary
\Dataset{}'s elements and that of the root must be calculated---this is called
the ``visit ratio'', $V_i$, %
and represents the mean number of completions at \Dataset{} $i$ for each
completion from the pipeline.
To calculate $V_i$,
we start with the pipeline's root ``visit ratio'' $V_0:=1$.
Then, we apply the following recurrence: $V_i=(C_i/C_{i-1}) \times (C_{i-1}/C_0)$,
where $C_i$ is the average number of items of work completed at \Dataset{} operation $i$.
The former ratio is the \Dataset's local input-output ratio, and the latter is
calculated in the recurrence.
Intuitively, the visit ratio allows one to say that $n$ elements are in a batch,
and thus dependencies to batching must have a throughput $n$ times faster to
``keep up''.

\textbf{Throughput Cost.}
The throughput at the root, $X_0$, is the number of minibatches completed,
$C_0$, in a timeframe, $T$, and the child \Datasets{} have $X_i=V_i X_0$.
However, this equation does not explain the bottleneck cause, which
requires reparameterizing the throughput in terms of CPU core
time or I/O bytes.
As $X_i=C_i/T$, we factor the equation into
1) the product of completions per resource (e.g., elements per
core-seconds) and
2) resource per
time (core-seconds per time).
The former is the ratio of two traced variables (element completions and
CPU-time or bytes used), and the latter is a knob for modeling adding or removing
resources (e.g., CPU parallelism or extra bandwidth).
In the LP (\S\ref{sec:linear_programming}), $R_i$ is the first factor normalized
by $V_i$, and $\theta_i$
is the second factor; in a bottleneck, they determine $X_0$.

\textbf{Materialization Cost.}
Estimating the size of a \Dataset{}'s materialized artifacts is similar to the
prior operational treatment but involves propagating the estimates \textit{up}
from data source to root.
The materialized size of a data source is the product of 1) the number of elements
(cardinality) and 2) the average size of each element.
Both are necessary because a \Dataset's semantics may modify one or the
other; for example, truncation only modifies the former and decompression only
the latter.
To start, the size of a data source is the number of files, $n$, times the average
bytes per file, $\bar{b}$.
Propagating the number of elements, $n_i$, involves multiplying $n$ by an input-output
completion
ratio.
The sum of output bytes and the number of completions for each \Dataset{} is measured in tracing, and thus
$\bar{b}_i$ is readily computed.
$n_i$ can grow unbounded (and thus uncacheable) if the data is infinitely
repeated or augmented.
\Datasets{} that are children to a cache can be modeled as having no
steady-state cost (e.g., after the first epoch).

\section{Evaluation}%
\label{sec:evaluation}%
We evaluate CPU bottleneck removal in \S\ref{sec:cpu_bottleneck}, showing that
\plumber{} can indeed find bottlenecks, and we further analyze how \plumber's solutions differ from those of baselines.
We additionally evaluate disk and caching in \S\ref{sec:disk_micro} and
\S\ref{sec:cache_micro}.
End-to-end results are presented in \S\ref{sec:end_to_end}.
The appendix provides additional details and demonstrates that \plumber's
overhead on modern hardware is less than 21\% (due to text workloads) and drops to below
5\% on vision workloads.
We compare against a naive configuration, which has minimal parallelism, to two
strong baselines: \autotune{}~\cite{murray2021tfdata} and
\heuristic{}, which set the parallelism tunables to the number of
cores on the machine.

\textbf{Hardware.}
For microbenchmarks, we evaluate over two setups to ensure our results generalize.
Setup A is a consumer-grade AMD 2700X CPU with 16 cores and 32GiB RAM\@.
Setup B is an older enterprise-grade 32--core Intel E5--2698Bv3 Xeon 2GHz CPU
with 64GiB RAM\@.
Setup C is for end-to-end results and is a TPUv3-8~\cite{jouppi2020domain} with
96 Intel Xeon cores and 300GB of RAM\@.

\textbf{Workloads.}
Our evaluation uses the MLPerfv0.6 subset of MLPerf
training~\cite{mattson2019mlperf} benchmarks, which is representative of
datacenter workloads and covers both images and text.
We use the following tasks and datasets:
\textbf{ResNet/ImageNet}~\cite{he2016deep,deng2009imagenet}, \textbf{Mask
RCNN/COCO}~\cite{ren2016faster,lin2014microsoft},
\textbf{MultiBoxSSD/COCO}~\cite{liu2016ssd},
\textbf{Transformer/WMT}~\cite{vaswani2017attention,WMT17}, and
\textbf{GNMT/WMT}~\cite{wu2016google,WMT16}.

\subsection{CPU Bottleneck Removal}%
\label{sec:cpu_bottleneck}%
To assess how accurately bottlenecks are found, we use \plumber's analysis layer
to rank nodes by bottleneck.
The pipeline's parallelism parameters are initialized to the naive configuration
(parallelism=1) with prefetching, and \plumber{} iteratively (using 1 minute of
tracing) picks the node to
optimize by ranking nodes by their parallelism-scaled rates.
To compare against uninformed debugging, we plot a random walk, which randomly
picks a node to parallelize for each ``step'' (x-axis in our plots).
We run each experiment three times to get confidence intervals.

\begin{figure}
  \begin{subfigure}[t]{0.50\linewidth}
    \leftExpResult{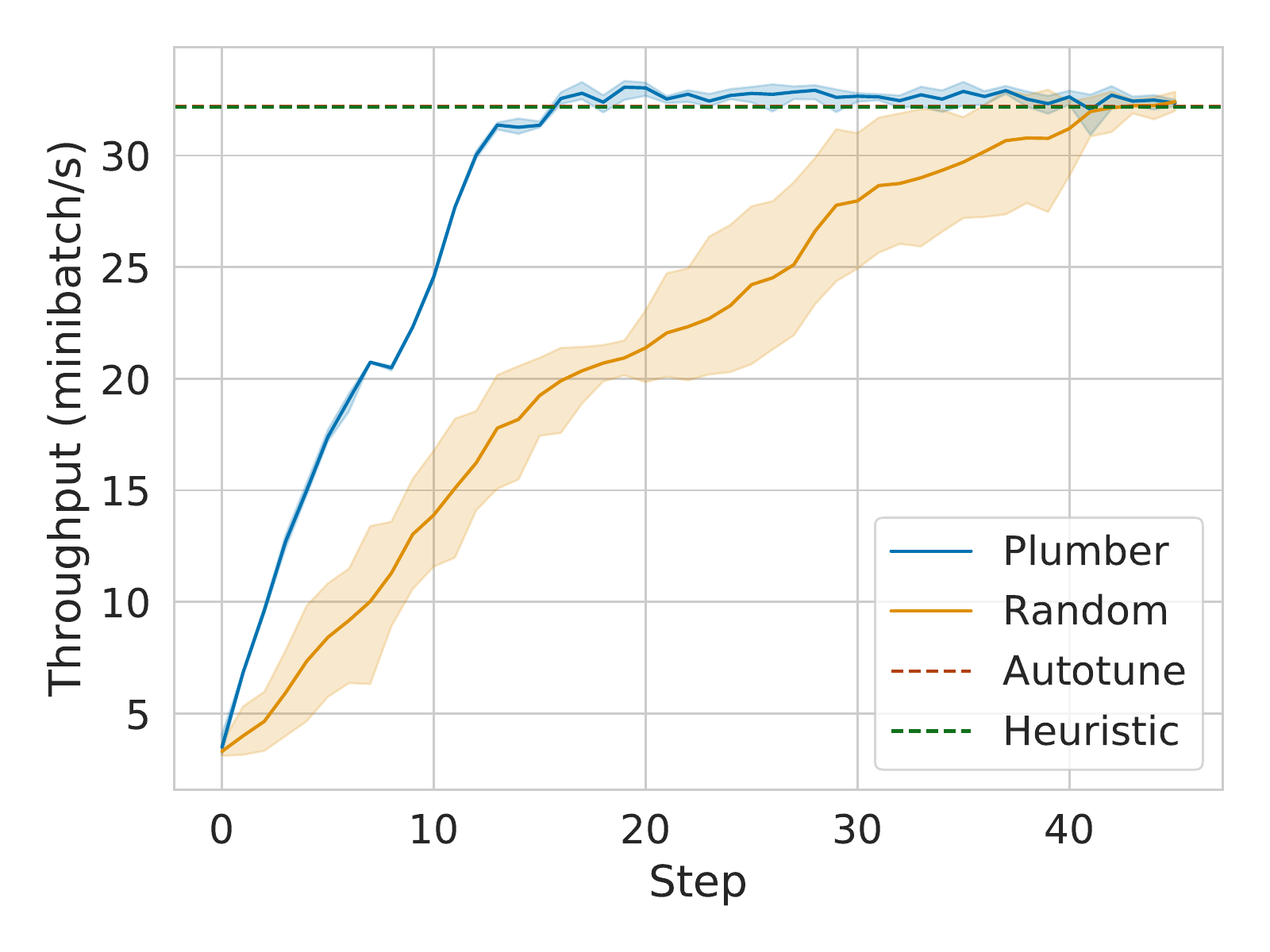}
    \caption{ResNet (Setup A)}
  \end{subfigure}%
  \begin{subfigure}[t]{0.5\linewidth}
    \rightExpResult{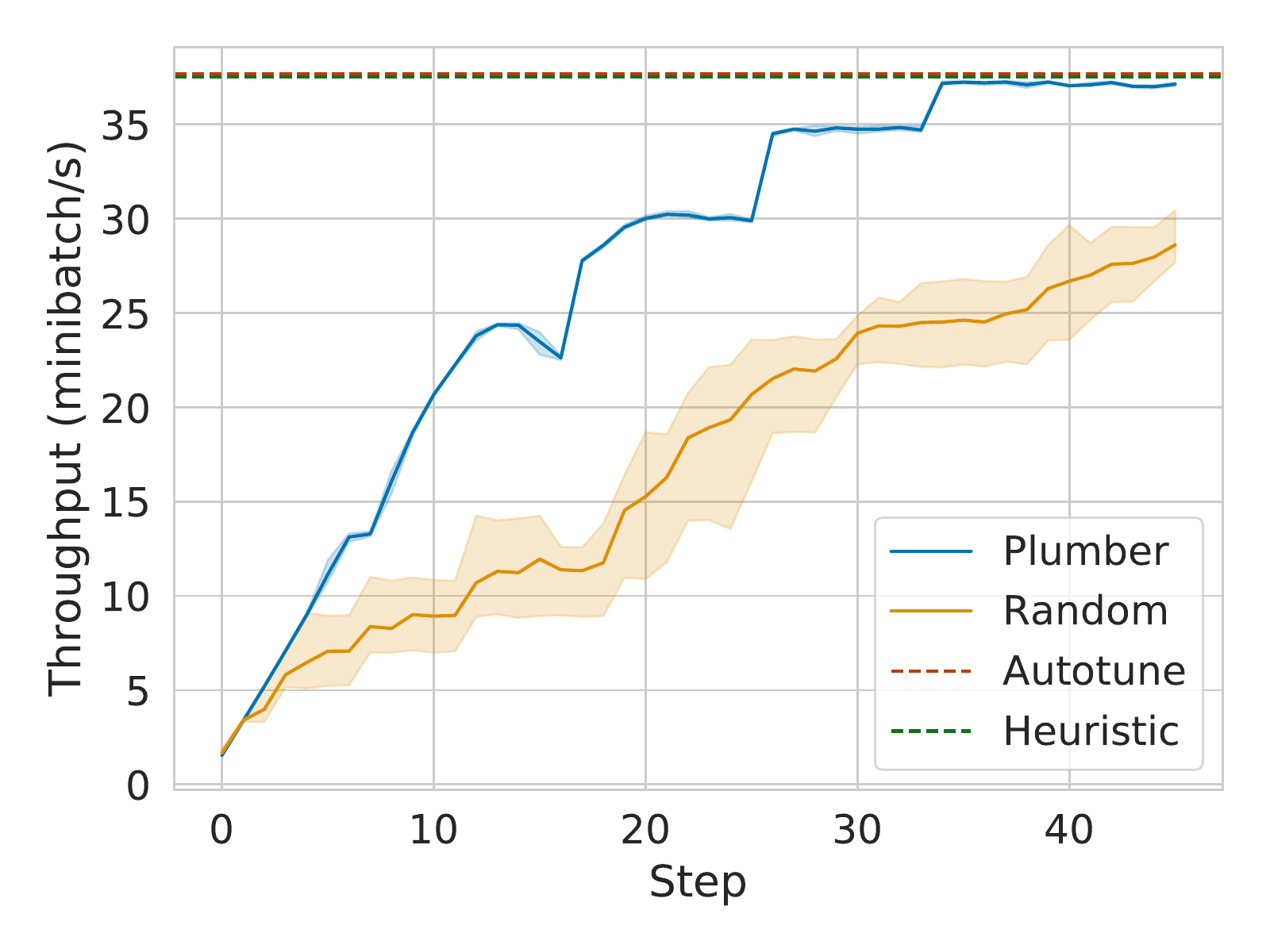}
    \caption{ResNet (Setup B)}
  \end{subfigure}%
  \vspace{-5pt}
  \caption{\plumber{} outperforms random walks by 2--3$\times$, demonstrating
  that \plumber's signal is markedly better than guessing.
  X-axis denotes the optimization step, starting at minimal parallelism.
  Y-axis denotes the pipeline rate in minibatches per second with 95\%
  confidence intervals.
  }%
  \label{fig:imgnet_convergence}
\end{figure}

\textbf{Sequential Tuning.}
Figure~\ref{fig:imgnet_convergence} shows the ResNet workload
across both setups; other workloads look similar.
\plumber{} is consistently better than the random walk, as expected.
We observe that both \heuristic{} and \autotune{} are equivalent in terms
of reaching peak performance---over-allocation does not usually result in
performance degradation.
On the ResNet workload, the bulk of the work is in the JPEG
decoding \Dataset{}, which services 2.5 minibatches/second/core on Setup
A. Most of Setup A's steps are spent increasing the parallelism of this
\Dataset{}, a transpose operation being the second bottleneck.
The bumps in Setup B correspond to increasing the parallelism of Transpose rather than JPEG decoding and occur roughly once every 8 steps.
We observe that, across all pipelines, such transition regions are the only regions (in addition to fluctuations at convergence) where \plumber{} struggles to characterize the locally optimal decision (see
MultiBoxSSD example in Appendix).
While Setup B has $2\times$ more cores than A, the per-core decoding rates
for B are lower, resulting in only a $1.2\times$ higher throughput.

\observation{\textit{\plumber{}'s bottleneck finder converges to the optimal
throughput in 2--3$\times$ fewer steps than a random walk.
Inspecting the individual \Dataset{} rates offers pipeline and machine performance
insights.}}

\begin{figure}
  \begin{subfigure}[t]{0.4975\linewidth}
    \leftExpResult{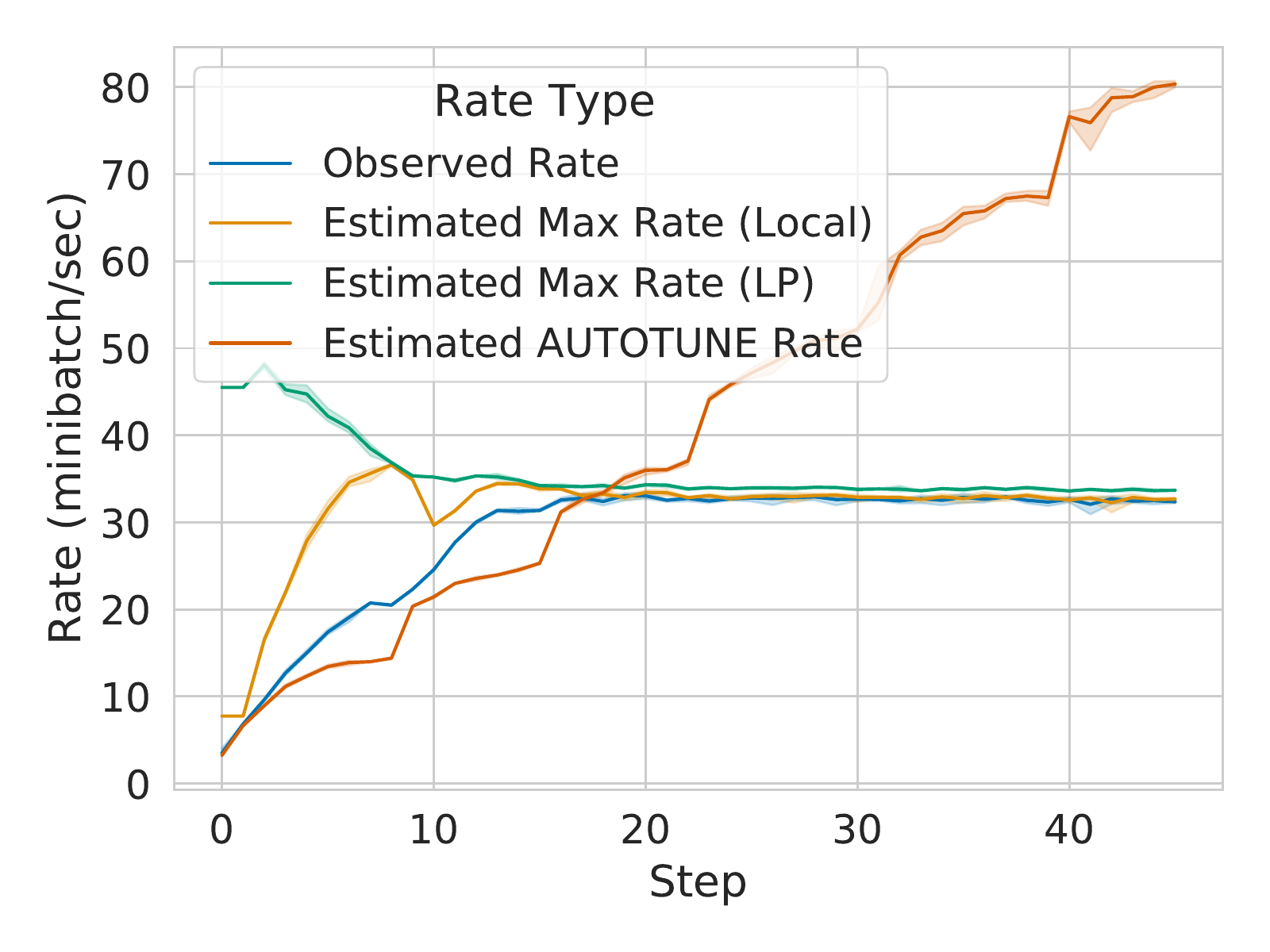}
    \caption{ResNet (Setup A)}
  \end{subfigure}%
  \begin{subfigure}[t]{0.5025\linewidth}
    \rightExpResult{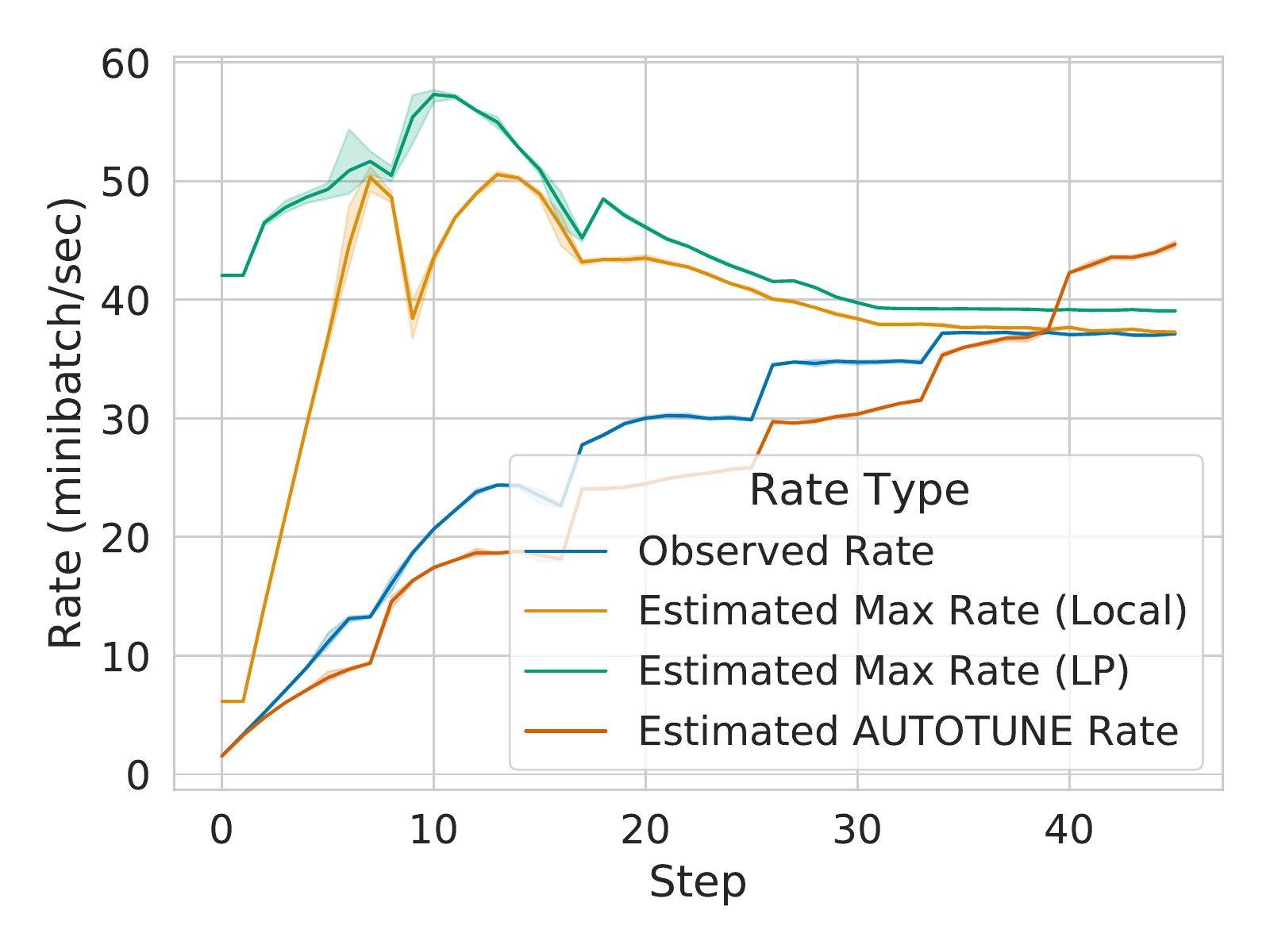}
    \caption{ResNet (Setup B)}
  \end{subfigure}%
  \vspace{-5pt}
  \caption{
  Before optimizations begin, \plumber{} is able to bound performance within
  $2\times$ with the LP, and the gap decreases over time.
  Setup B exhibits superlinear scaling around step 10 and also
  exhibits more pronounced bottlenecks, as it takes longer to converge.
  The \autotune{} model does not account for saturation and therefore
  has unbounded predicted throughput.
  }%
  \label{fig:imgnet_convergence_LP}
\end{figure}

\textbf{Linear Programming.}
Figure~\ref{fig:imgnet_convergence_LP} demonstrates that \plumber{} can
understand performance through the LP formulation on ResNet; other workloads are similar.
As a baseline, we include a ``local'' method, which allocates all remaining
resources to the current
bottleneck node.
This baseline is unable to see past one bottleneck and thus oscillates as the
bottleneck node changes (the ``bumps'' in Figure~\ref{fig:imgnet_convergence}).
Meanwhile, the LP steadily declines---upon inspecting the solution, we find a
strong correlation with the LP's decline and the value of $R_i$ for the
bottleneck node (JPEG decoding).
While $R_i$ typically decreases (e.g., due to scaling overhead), we find it
\textit{increases} briefly especially on Setup B, peaking at step 10 and resulting in a
``bell shape'' LP prediction (JPEG rate peaks at 1.8 and then drops to 1.4 by the end of training),
explaining the peak in the LP\@. 
\autotune{}, being oblivious to saturation, either overestimates or underestimates
throughput without bounding it.

\observation{\textit{\plumber's LP solution captures \textit{both} resource
utilization and bottlenecks, bounding throughput to within $2\times$ from when
optimization \textit{starts} for pipelines like ResNet and MultiBoxSSD\@.
The bounds get tighter as optimization proceeds due to differences in empirical rates.}}

\begin{figure}
  \begin{subfigure}[t]{0.50\linewidth}
    \leftExpResult{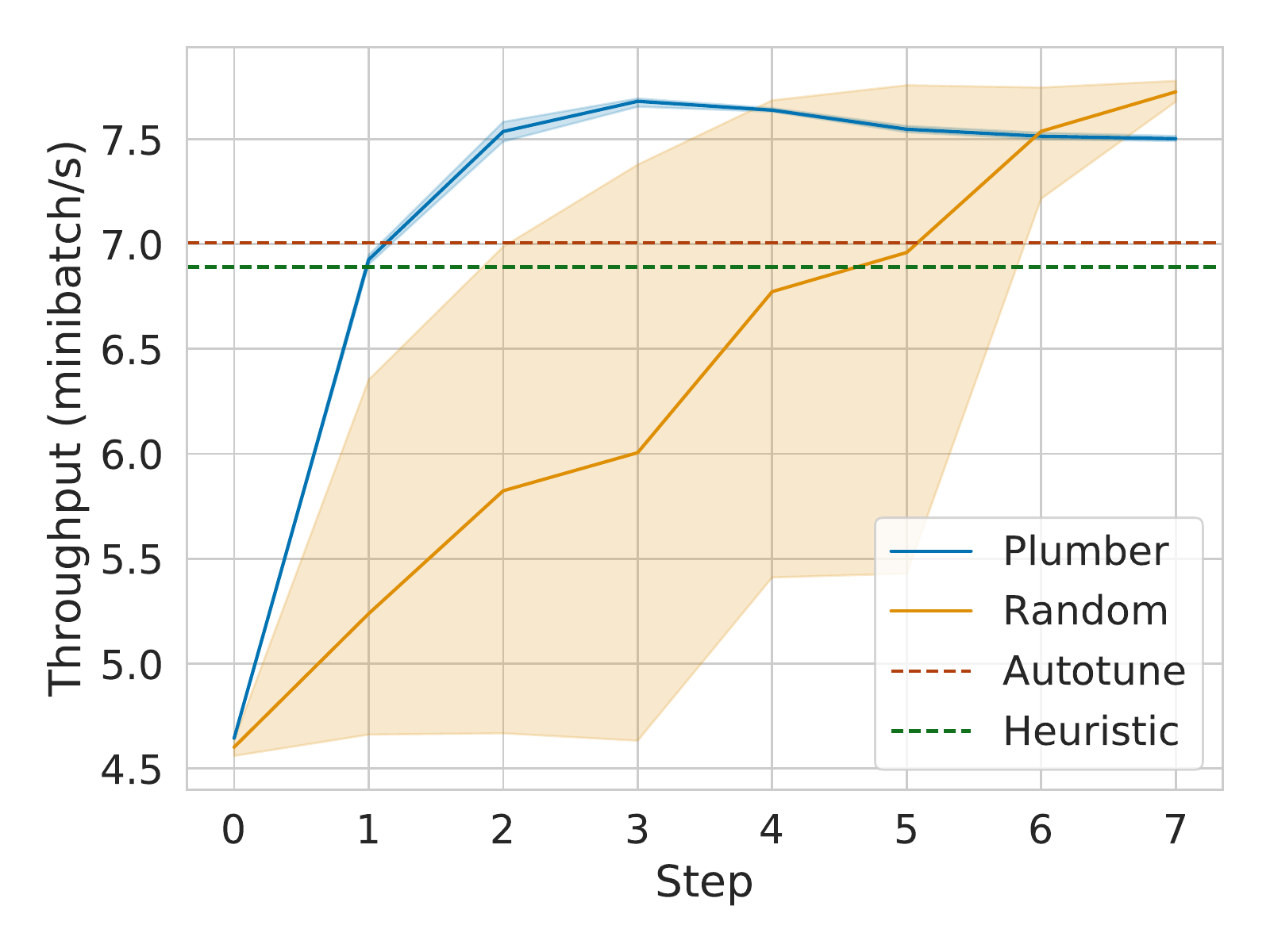}
    \caption{RCNN Convergence}
  \end{subfigure}%
  \begin{subfigure}[t]{0.50\linewidth}
    \rightExpResult{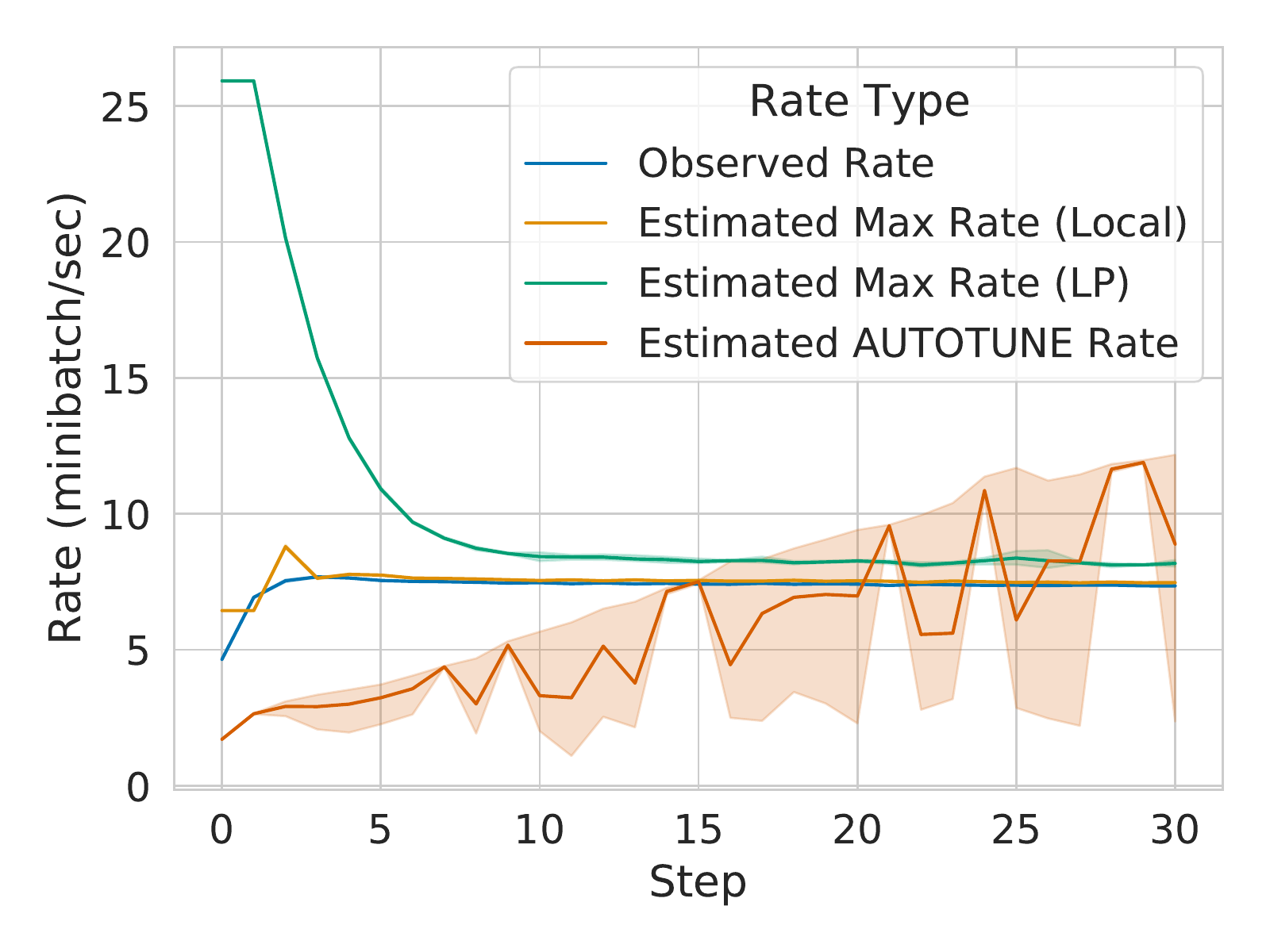}
    \caption{RCNN Predictions}
  \end{subfigure}%
  \vspace{-5pt}
  \caption{
    RCNN on Setup A, along with its predictions.
    RCNN exhibits heavy UDF parallelism, which causes thread over-allocation to quickly deteriorate performance.
    \autotune{} has high variance estimates of latency.
  }%
  \vspace{-15pt}
  \label{fig:rcnn_convergence}
\end{figure}

\textbf{Large UDF Parallelism Challenges.}
As shown in Figure~\ref{fig:rcnn_convergence}, RCNN on Setup A is challenging
and displays counter-intuitive behavior.
The reason is that RCNN features a large UDF in its \texttt{MapDataset},
which is transparently parallelized by the \TensorFlow{} runtime.
Parallelizing the \texttt{Map} is dangerous because the parallelism compounds
with UDF parallelism---1 parallelism uses nearly 3 cores!
As parallelism increases for this \Dataset{}, the corresponding per-core rate
drops, causing the LP prediction to drop and baselines to overshoot peak
performance on both setups.
While the LP overestimates peak
performance by $4\times$, it is still qualitatively better than
\autotune{}, which oscillates in predictions.
Upon inspection, \autotune{} allocates 16 parallelism to the
bottleneck node, but 3 parallelism to a different \texttt{MapDataset} node.
The bottleneck node operates at 0.5 minibatches/second/core, while the other
node operates at 20 minibatches/second/core.
Thus, the optimal policy, which \plumber{} follows, is to only allocate
parallelism to the main bottleneck (thus bounded by $0.5 * 16=8$).
In fact, due to UDF parallelism, only 4--5 parallelism is necessary.
Counterintuitively, this policy is no longer optimal for our end-to-end results
(\S\ref{sec:end_to_end}), which have $6\times$ more cores.

\observation{\textit{\autotune{} and \heuristic{} are vulnerable to over-allocation, which can cause performance degradation.
Pipelines with heavy UDF parallelism may experience drops on the order of 10\%.
Dynamic parallelism makes end-to-end performance hard to predict.}}

\begin{figure}
  \begin{subfigure}[t]{0.5\linewidth}
    \leftExpResult{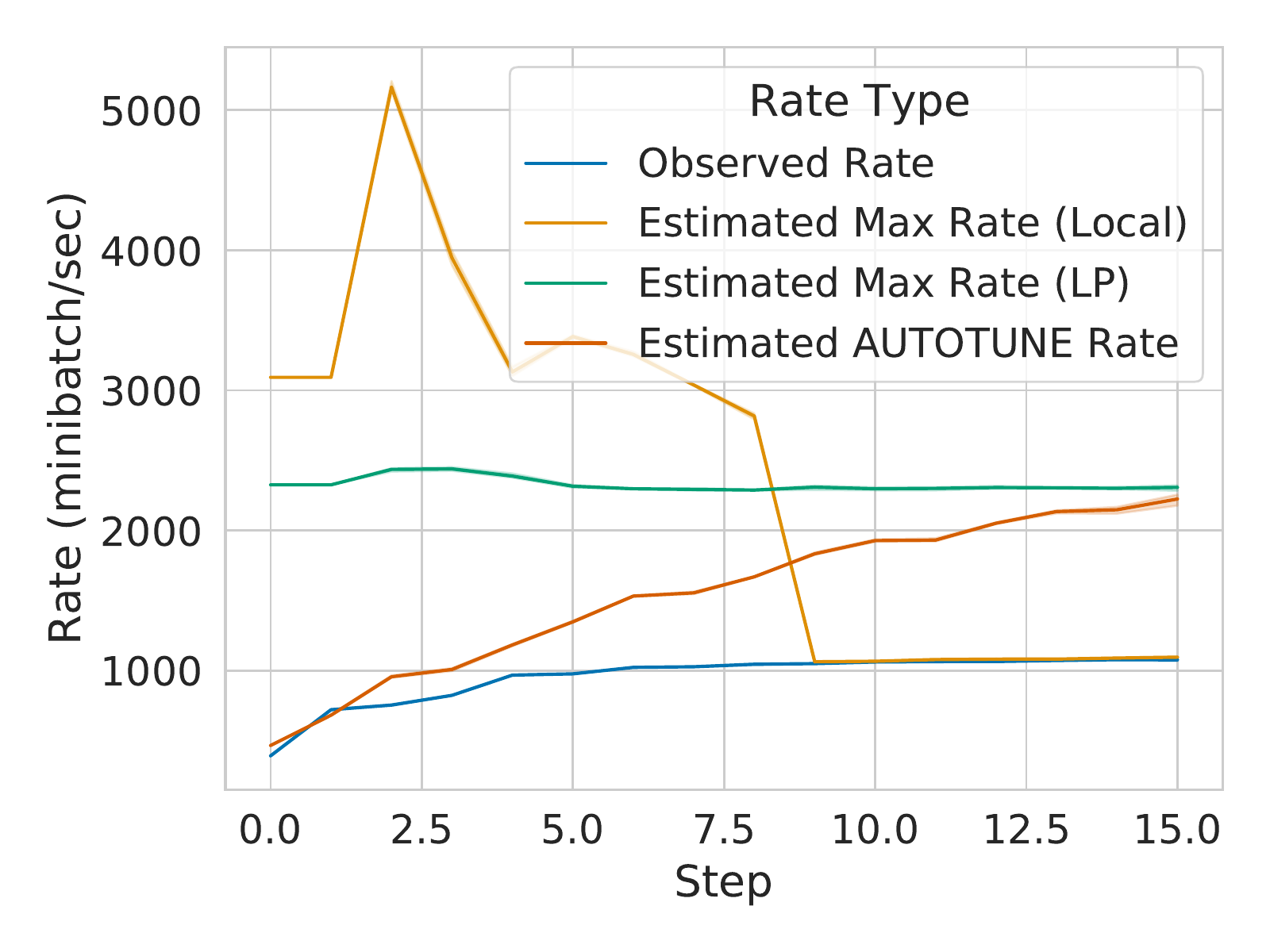}
    \caption{Transformer Predictions}
  \end{subfigure}%
  \begin{subfigure}[t]{0.5\linewidth}
    \rightExpResult{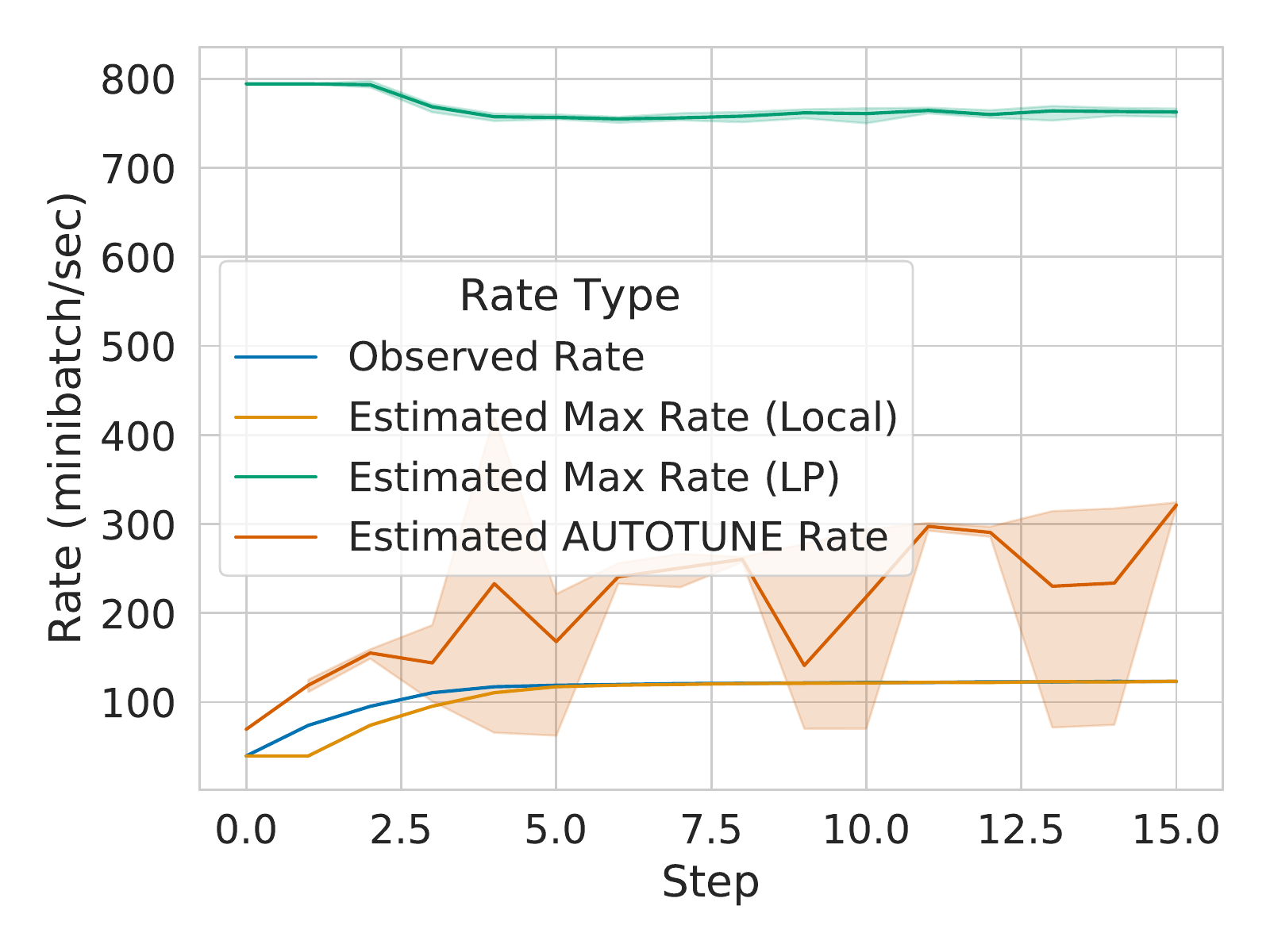}
    \caption{GNMT Predictions}
  \end{subfigure}%
  \vspace{-5pt}
  \caption{
    Transformer and GNMT predictions on Setup A.
    Transformer and GNMT exhibit small operations, which are a mismatch to the \Iterator{} model.
    It is difficult to fully saturate a CPU with \Dataset{} parallelism alone---caching or outer parallelism are required.
  }%
  \vspace{-15pt}
  \label{fig:text_convergence}
\end{figure}

\textbf{Text Processing (NLP) Challenges.}
We observe that both Transformer and GNMT are difficult to optimize in practice.
As shown in Figure~\ref{fig:text_convergence}, both pipelines are predicted to
be 2--8$\times$ faster than they actually end up being.
Upon investigation, we observe that nearly all operations in NLP are very small e.g., grouping a few integers in a vector.
The operations are so small that they are significant compared to the \Iterator{}
abstraction's
overhead, causing idle ``bubbles''.

According to \plumber{}, GNMT is bottlenecked by
\texttt{Shuffle\-And\-Repeat\-Dataset}; this \Dataset{}
is performing minimal work and thus the result is unexpected.
Before getting to this point, \plumber{} allocated 9 parallelism to the first
\texttt{MapDataset} and then gave up upon seeing the non-optimizable \Dataset.
Similarly, for Transformer, after alternating optimizing the 3 \texttt{MapDatasets} (all similarly fast),
\plumber{} indicates Transformer is bottlenecked by \texttt{FilterDataset},
which is operating at about half of its
max rate (explaining the $2\times$ difference).
Using outer-parallelism for both pipelines, introducing inner-parallelism for
\texttt{Batching} (GNMT), and resuming \plumber's optimization results in
closing the predicted performance gap to less than $2\times$ (Transformer) and $4\times$ (GNMT).

\observation{\textit{Text pipelines can require significant tuning to overcome framework overheads. An in-memory cache materializing results is ideal, when possible.}}

\subsection{Disk Microbenchmarks}%
\label{sec:disk_micro}%
To simulate various bandwidths, we implement a token-bucket bandwidth limiter
in the \TensorFlow{} filesystem layer and use the most I/O intensive
pipeline, ResNet.
\plumber{} correctly concludes that ImageNet records are approximately 110KB on average
and infers that 128 records are necessary for 1
minibatch.
It thus infers that the I/O load per minibatch is $128*110$ KB, or 6.9
minibatches per 100MB/s of bandwidth, which it can join with known bandwidth (e.g., from token-bucket or \texttt{fio}~\cite{fio}).
Using this bound, \plumber{} predicts the pipeline's performance within $5\%$
from 50MB/s to 300MB/s, when the compute bottleneck begins. We observe similar
results for RCNN and MultiBoxSSD, though MultiBoxSSD is easier to bottleneck consistently due to its faster CPU speed.
\plumber{} estimates that RCNN and MultiBoxSSD can push 145 minibatches/sec on 100MB/s, since they use the same
dataset and same batch size; therefore, MultiBoxSSD is $25\times$ more I/O bound
for a fixed CPU\@.

We then run this experiment on setup B with a real HDD (Seagate ST4000NM0023)
and NVMe SSD (400 GB Intel P3600), which have 180MB/s and 2GB/s of read bandwidth,
respectively.
We run the load for one minute to prevent the page cache from kicking in (when the
dataset reads repeat).
For ResNet, \plumber{} predicts 11.06 minibatches/sec, and we are bound at 12.7 (15\% error).
On the NVMe SSD, \plumber{} predicts 135 minibatches/sec, and, indeed, we observe a compute
bound.
On RCNN, \plumber{} predicts disk bounds of 970 and 11850 minibatches/sec, respectively; for
both, we observe the compute bound of 14 minibatches/sec.
On HDD with MultiBoxSSD, \plumber{} predicts 235 minibatches/sec, whereas we observe 215 (10\% error).
On NVMe SSD with MultiBoxSSD, \plumber{} predicts a disk bound of 2900
minibatches/sec;
we observe the compute bound being hit before the drive is saturated.
The text datasets are too small to test.

\observation{\textit{\plumber{} is able to bound disk-bound workloads to within
15\% of the observed throughput, notifying users of potential hardware misconfigurations.}}

\subsection{Memory (Cache) Microbenchmarks}%
\label{sec:cache_micro}%
We evaluate \plumber{}'s predictions on memory optimization and summarize them
below.
\plumber{} predicts that 148GB are necessary to cache ImageNet for ResNet; 20GB to cache
COCO for RCNN and MultiBoxSSD; and 1GB and 2GB for the Transformer and GNMT WMT
datasets, respectively, which matches their known size.
While this is expected for a full sweep of the training set (by simply tracking
all file sizes), it also holds for sampling the dataset.
Empirically, we find that observing a small subset of the data is sufficient:
1\% of files provide a relative error of 1\% for ImageNet and 2\% for COCO, and
5 files gives less than 2\% relative error for WMT datasets.

For materialized caches, we observe that \plumber{} predicts no changes until a
node is hit, which changes the bytes/element.
For ImageNet, we observe that, if \plumber{} is fed a fused decode and crop
pipeline, it predicts caching is only possible at the source, because the crop
is random.
However, when the ImageNet pipeline is not fused, image decoding amplifies the dataset size
by $6\times$, which \plumber{} observes as 793GB of a true 842GB, or 6\%
error with 60 seconds of profiling.
For this pipeline, we observe that relative error decreases as a function of tracing
time, saturating at 2 minutes and yielding relative error rates below 1\%,
offering a knob for refining estimates at the expense of tuning time.
RCNN can only be cached at the disk-level, since the following UDF is randomized.
For MultiBoxSSD, \plumber{} detects it takes 84GB of a true 97GB (14\% error) to materialize the dataset in memory after
image decoding with only 60 seconds of profiling, with 2 minutes yielding a 5\%
error, which continues to drop by $\sim$1\% for each additional minute.
\plumber{} additionally detects that the filter in the pipeline reduces the dataset by less than 1\%.

\observation{\textit{\plumber{} captures dataset sizes at the source exactly,
and, for large datasets, it is able to subsample 1\% of files to obtain 1\% error. For materialized caching, \plumber{} propagates changes to dataset sizes (e.g., data decompression and filtering).}}

\subsection{End-to-End Pipeline Optimization}%
\begin{figure}
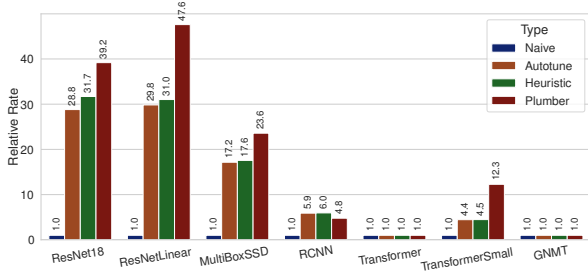
%
\centering%
\resizebox {0.99\linewidth} {!} {
\begingroup%
\makeatletter%
%
\makeatother%
\endgroup%
 }
  \vspace{-5pt}
  \caption{
    Relative speedups over the naive configuration for end-to-end model training on
    TPUv3-8.
    Apart from RCNN, \plumber{} surpasses strong baselines by adding caching, yielding speedups
    of up to $47\times$ compared to naive and 50\% compared to tuners.
    While text pipelines are challenging to improve on MLPerf, using a smaller Transformer
    combined with a more complex pipeline can increase the tuners gap to $2.5\times$.
  }%
  \label{fig:gcp_all}%
  \vspace{-20pt}
\end{figure}

\label{sec:end_to_end}%
The end-to-end benefits of \plumber's optimizations, evaluated over 5 epochs of training,
are shown
in Figure~\ref{fig:gcp_all}.
None of the pipelines have caches inserted manually, naive configurations have 1
parallelism and no prefetching, and \heuristic{} uses the prefetching
hard-coded into the dataset.
We observe overprovisioning (\heuristic{}) is competitive, if not faster,
than \autotune{} across all trials.
\plumber{} can go beyond both of these strong baselines because of caching,
obtaining up to a $47\times$ speedup.
Such a speedup, an absolute throughput of over 14k images/second, is only possible because caching bypasses the
(\plumber{}-derived) 11k image/second data source bottleneck for the cloud
storage.
For ResNet-18, it is sufficient to cache at the data source, which is only
148GB; therefore, \plumber{} picks the CPU-optimized branch of the pipeline
(code in Appendix).
However, when we use a linear model for ResNet, we use the smaller validation set, which allows
\plumber{} to cache the $6\times$ bigger decoded images in memory, avoiding a
CPU bottleneck.
We also evaluate ResNet-50 and find that \plumber{} obtains a $24\times$ speedup
over the naive configuration, though it cannot improve over other baselines, as
the model's throughput limits are hit at 8k images/second.

Compared to \autotune{}, \plumber{} obtains a 36--59\% speedup on three of the MLPerf
benchmarks and ties on Transformer and GNMT (due to model throughput).
These numbers are conservative---we observe that \autotune{} on ResNetLinear often sets the I/O parallelism to 1,
which results in a 35\% performance drop compared to what is shown
in Figure~\ref{fig:gcp_all} and creates a
$2.4\times$ performance gap between \autotune{} and \plumber{}.
To allow more competitive tuning from \autotune{} on that workload, we set the I/O parallelism to the default value
of 10 used in MLPerf submissions, which improves its final performance.
For RCNN, \plumber{} may perform worse than \autotune{} because \plumber{}
is conservative in its parallelism allocation, while \autotune{} tends
to allocate maximum parallelism to all \Datasets.
\plumber{} estimates that one \texttt{MapDataset} is two orders of
magnitude more expensive than the other.
In some cases, \plumber{} allocates 95 parallelism to the former,
leaving only 1 parallelism for the remaining \texttt{MapDataset}, which
results in the shown 23\% performance loss.
However, we also observe cases where \plumber{} matches the throughput of the
baselines because it allocates at least 2 parallelism to the cheaper \Dataset{},
which removes the bottleneck---suggesting that a mild form of ``hedging'' would
be effective in preventing under-allocation for such skewed workloads.
For MultiBoxSSD, \plumber{} is able to materialize the data after
filtering is performed, which makes the cache smaller and increases throughput
by removing load from the CPU\@.
While we don't see a large benefit from \plumber{}'s optimizations on the MLPerf
NLP pipelines, we do see improvements when moving to the related official
\texttt{Flax} implementations of Transformer when it is configured to use a single-layer Transformer (TransformerSmall).
We see a $2.5\times$ difference between strong baselines and \plumber{} on
TransformerSmall because the \texttt{Flax} pipeline performs text-processing and packing on-the-fly, which, when combined with
a smaller model, makes peak throughput only achievable with aggressive caching.

\observation{\textit{\plumber{} frees the user from making individual caching
and parallelization decisions, enabling $50\%$ end-to-end improvements over
\autotune{} and heuristics and $40\times$ improvements over naive configurations.}
\section{Related Work}%
\label{sec:related_work}%

\textbf{Pipeline Optimization.}
Dataset Echoing~\cite{data-echoing} repeats input pipeline operations to match
compute rate.
DS-Analyzer predicts how much file cache memory is necessary to match the compute steps~\cite{dsanalyzer}.
Progressive Compressed Records~\cite{PCR} match compression levels to likewise minimize I/O.
Each of these works is characterizing a piece of the input pipeline, all of which can be homogenously dealt with via \plumber{}---the first is a visit ratio and the latter two are memory caching/disk.

Systems have also been built to natively support the data pipeline workload.
A cache library has been developed to carefully partition and coordinate
caches in distributed training ~\cite{dsanalyzer}
Dataset echoing has been further expanded to support caching partially augmented
samples~\cite{refurb_data}.
Filesystem middleware has been developed to
natively support data prefetching~\cite{clairvoyant_prefetching}.
While \plumber{} primarily focuses on native \tfdata{} primitives, it could
also be used to hook into such systems to perform cross-system optimization.

\textbf{Bottleneck Detection.}
Tools in the big-data domain (e.g., Spark \cite{zaharia2012resilient}) have similar performance problems as those found in ML pipelines \cite{ousterhout2015making,ousterhout2017monotasks}, but the differences in domain encourage a different approach.
Notably, \texttt{Monotasks}~\cite{ousterhout2017monotasks} enables
bottleneck finding by designing Spark primitives to be easily
measured on a per-resource level.
In contrast, \plumber{} does not modify the framework and instead elects to
carefully instrument selected resource usage, though the design is similarly
simplified by having resource-specialized operations.
Big data systems, such as DyradLINQ~\cite{fetterlydryadlinq}, have similarly benefited from dedicated
debugging utilities~\cite{daphnedryad} and dynamic query rewriting~\cite{optimus}.
Roofline~\cite{williams2009roofline,extended_roofline}
models bound compute kernels
with CPU limits. \plumber{} generates similar plots using \Dataset{} and resource limits.

Recent studies have analyzed ML workloads and found that data
pipelines can become bottlenecks~\cite{dsanalyzer,murray2021tfdata}.
This line of work is further supported by empirical evidence of
bottlenecks in recommendation models~\cite{zhao2021understanding}.
Recent work has also characterized the design space of input
pipelines with a profiling tool~\cite{bottleneck_sigmod}; we do not focus on
general pipeline decisions (e.g., the choice of image resolution), but rather
focus on automatic tuning of performance knobs.
These works all highlight a growing need for systems which better support the
input pipeline workload.

\section{Conclusion}%
\label{sec:conclusion}%
Training ML models requires reading and transforming
large quantities of unstructured data.
Our analysis of real ML training jobs at \Google{} shows that data pipelines do not adequately utilize their
allocated hardware resources,
leading to potential bottlenecks, which waste precious accelerator resources.
To enable practitioners to better utilize their hardware and improve training performance,
this paper introduces \plumber, an extensible tracing and optimizing tool.
\plumber{} localizes bottlenecks to an individual operator by
tracing both the resource usage and rate of each operator, enabling both
bottleneck-finding as well as inferences on future performance.
By adding caching in addition to tuning, \plumber{} can surpass the end-to-end
performance of existing tuners by 50\%.
Future extensions to \plumber{} include:
distributed and accelerator-infeed level profiling,
optimal resource allocations for pipelines,
and
semantic-level re-writing of augmentations.

\section*{Acknowledgements}
We thank the anonymous reviewers for their help improving the presentation of this paper.
This material is based upon work supported by the U.S. Army Research Office and the U.S. Army Futures Command under Contract No. W911NF-20-D-0002.
The content of the information does not necessarily reflect the position or the policy of the government and no official endorsement should be inferred.
We thank the members and companies of the PDL Consortium: Amazon, Google, Hewlett Packard Enterprise, Hitachi Ltd., IBM Research, Intel Corporation, Meta, Microsoft Research, NetApp, Inc., Oracle Corporation, Pure Storage, Salesforce, Samsung Semiconductor Inc., Seagate Technology, Two Sigma, and Western Digital for their interest, insights, feedback, and support. 
Michael Kuchnik was supported by a National Defense Science and Engineering Graduate Fellowship.
This research was supported with Cloud TPUs from Google's TPU Research Cloud and research credits from Google Cloud Platform.

\clearpage

\bibliography{ms}
\bibliographystyle{mlsys2022}

\clearpage
\appendix
\section{Resource Accounted Rates Algorithm (Extended)}%
\label{sec:resource_accounted_rates_long}%

The LP formulation in the main text hinges on our ability to predict two \textit{resource accounted rates}: \circled{1} the rate $R_i$ of a \Dataset{} given its configuration (e.g., number of cores assigned to each operation) and how that rate would change when one operation is assigned more cores, %
and \circled{2} the number of bytes needed to cache a \Dataset{} at a given operation, accounting for bytes added/removed by data transformations applied by prior operations.

The intuition behind the computation of \circled{1} is that the rate of each operation from the source (e.g., data read from disk into the pipeline) to the root (i.e., minibatches exiting the pipeline) differs. For example, map operations may filter or amplify their inputs. We therefore begin with the root of the \Dataset{} tree and traverse the pipeline until the source, accounting for every operation's rate in the process. The intuition behind the computation of \circled{2} is that the total data that needs to be cached is known at the source (i.e., when it enters the pipeline) and needs to be recomputed after each operation is applied. We therefore begin with the source of the \Dataset{} tree and traverse the pipeline until the root.

\textbf{\circled{1} Work Completion Rates.}
In input pipelines, each \Dataset{} operation is potentially characterized by different units of work.
Looking at Figure~\ref{fig:tfdata_graph}, we do not know how many bytes are in a minibatch or how \texttt{Map} operations turn bytes into training examples, but we can observe that 128 images are batched into a minibatch.
We thus rely on \textit{visit ratios} from Operational Analysis~\cite{denning1978operational}, i.e., normalization constants, to compute the equivalent of root units, i.e., minibatches per second per core, that are generated by the source.
Each \Dataset{} operation is characterized by its local rate, $r_i$, which is defined as the average items of work completed per second per core, and can be computed by tracking two items: the items arrived and the items completed.
The visit ratio, $V_i$, then, for a given \Dataset{} operation is the constant that converts the local rate, $r_i$, to the global rate, $R_i$, which expresses the average amount of work completed by the operation in minibatches per second per core, i.e., the unit of work of the \Dataset{} root given one core.
Starting with the pipeline's root visit ratio, $V_0$, which is equal to 1, \plumber{} computes each \Dataset{} operation's visit ratio using $V_i=(C_i/C_{i-1}) \times (C_{i-1}/C_0)=r_i \times V_{i-1}$, where $C_i$ is the average number of items of work completed at \Dataset{} operation $i$.

While the pipeline is running, \plumber{} collects counters on items arrived and
completed across all operations of a given \Dataset{}, which is used to model
CPU performance. This rests on two simplifying assumptions, which do not affect
\plumber{}'s efficiency in practice (\S\ref{sec:evaluation}): CPU work per item
is fixed, and there is linear scaling of operation performance.
Extending this approach to measure disk bandwidth usage only requires measuring filesystem reads and dividing by the wallclock time of the process. Converting usage to utilization is a matter of dividing the bandwidth usage by the available bandwidth, which \plumber{} measures by profiling the training directory using \texttt{fio}~\cite{fio}.

\textbf{\circled{2} Cache Amplification Rates.}
In ML settings, the benefit of caching is heavily skewed toward caching the \emph{entire} disk-resident data---there is no locality due to random sampling.
The fastest caching solution is one that is closest to the root of the \Dataset, while remaining in memory.
We note that \textit{random} \Datasets{}, which call into randomized functions, do not qualify for caching. By virtue of being random, their effective size is infinite---often, one can continue generating unique results forever.

To determine memory requirements for caching, \plumber{} assumes that files are
consumed to completion and tracks the size of all input files (i.e., bytes read
until end of file) used during the pipeline's operation. Once a new file is
recorded, the file is added into a system-wide map tracking filename to bytes
used. The space requirements are moderate for realistic use-cases; only two
integers are needed per filename with hashing, and the number of files is likely small due to large files (e.g., ImageNet has 1024 files).

To calculate how many bytes it takes to materialize each \Dataset{}, we approximate two statistics: the number of elements (\textit{cardinality}), $n_i$, and the average size of each element (\textit{byte ratio}), $b_i$.
For a data source, we have the \Dataset{} size in bytes: $\sum_{f\in\mathbb{F}} S_{f}$, where $S_f$ is the size of a file $f$ in the set of recorded files, $\mathbb{F}$. As the source data flows through the input pipeline, $n_i$ and $b_i$ will change as training examples are grouped, filtered, and/or transformed (e.g., parsed and decoded in \texttt{Map}).
To build intuition, let us revisit Figure~\ref{fig:tfdata_graph}.
$b_i$ for \texttt{Map} is just the ``decompression ratio'' from records to
images, which is often approximated to be $10\times$ for JPEG\@.
Likewise, the \texttt{Batch} makes $b_i$ 128 times bigger in the process of grouping, while making $n_i$ 128 times smaller.
Therefore, we can conclude the root \Dataset{} is an order of magnitude larger than it was at the source.

We first tackle how to solve $n_i$.
For exposition, assume the common case that we only have one source (e.g., Figure~\ref{fig:tfdata_graph}) and the source is finite---we will be propagating our analysis up from this source until we hit the root.
Readers emit elements at per-row or per-record granularity, thus, for source
\Datasets, we overload $r_i$ from visit ratios to denote this ratio (though it is a ratio of
records/byte).
The initial dataset size at source $i$ is thus $n_i=(\sum_{f\in\mathbb{F}} S_{f}) \times r_i$.
For a \Dataset, $i$, the subsequent \Dataset{}, $j$, follows the
recurrence $n_j=r_j \times n_i$, where $r_j$ follows previously defined
semantics from visit ratios.
Intuitively, we convert the size in bytes to the size in \textit{records} to the size in training examples, and so on.
Special care must be taken to account for \Datasets{} repeating examples or
truncating the examples stream.
When generalizing to multiple children (multiple sources), some \Dataset-specific aggregation is required (e.g.,
\texttt{sum}).

To obtain $b_i$, we can use entirely local information.
\plumber{} tracks one local quantity, the bytes-per-element,
$b_i=\texttt{bytes\_produced(i)}/C_i$, where the numerator is a counter for
bytes produced at $i$, and $C_i$ is the counter for number of completions from $i$.
The accuracy of estimating the materialized size with $b_i \times n_i$ depends on the variance of each component's estimate.

To deal with large datasets, which may take longer than the \plumber{} tracing
time to iterate through, we obtain a subsample, $s$, of the set of file sizes, $S$, which
we can use to approximate $S$. If we have $n$ of $m$ samples, we can simply
rescale the subsampled size, $\sum_n{s}$, by $m/n$ to get an estimate of the full
dataset size, $\frac{m}{n} \times \E[\sum_n{s}]$.
Empirically, the
estimate is sufficiently accurate: 1\% of files gives relative error of 1\% ImageNet and 2\% for COCO, and 5 file gives less than 2\% relative error for WMT datasets.
This allows \plumber{} to get a tight estimate of the true dataset size in
seconds with sufficient read parallelism.
The normal distribution of error we observe matches intuition that the
Central Limit Theorem applies.
\section{Implementation Details}%
\label{sec:implementation_extra}%
\plumber{} is currently implemented with a fast but simple C++ core on top of
\tfdata{} (3k line patch) and a higher-level Python interface, analysis, and
graph rewriting code (8k lines).
\plumber{} builds on \tfdata{}'s C++ \autotune{} harness to collect statistics for each
pipeline \Iterator.
For each \Iterator{} operator, a struct captures the timer and counter-related
statistics, along with additional meta-data
totaling less than 144 bytes,
and the number of \Iterator{} operators is rarely over 1000.
We describe how these fields are collected and how they are used to derive
pipeline
statistics when \plumber{} ingests the output file (which is small---less than
200kB for ImageNet/ResNet).

\textbf{Measuring CPU.}
To measure CPU work, \plumber{} adds fine-grained CPU-timer support
to \TensorFlow{} and uses this
support to obtain more accurate statistics.
We use a similar architecture to \autotune{}, which starts and ends timers at the boundaries of \getnext{} calls.
There are two advantages to using CPU-timers.
First, the sum over \Dataset{} time closely tracks CPU utilization, yielding a
valid $\theta_i$, which in practice bounds the number of generated threads.
Preliminary tests showed that summing over wallclock time rather than CPU time can overestimate
ImageNet/ResNet pipeline's utilization by $2.5\times$, with error starting
to grow when cores are oversubscribed. %
Second, \Datasets{} that use significant wallclock time but not much CPU time are
accounted correctly---examples include \Datasets{} that sleep or read from files.
There is a cost associated with using this different timer. %
This overhead, in the order of tens to hundreds of nanoseconds, becomes visible when work is so small that the overhead is not amortized out, typically when CPU utilization is low.
While this price is only paid during the tracing stage, we can reduce the overhead via subsampling, adding vDSO plugins, or using wallclock timers.

\textbf{Graph Rewrites.}
Applying \texttt{Graph} (a \TensorFlow{} \tfdata{} program in our
application) rewrites requires a compiler-like approach.
Because a \texttt{Graph} is instantiable,
\plumber{} takes as input a serialized \texttt{Graph} representation of the input pipeline and outputs a modified \texttt{Graph} representation.
\plumber{} builds an in-memory representation of \Datasets{} to make
decisions, such as what node to parallelize or cache.
To connect the internal representations
with the \texttt{Graph}, \plumber{} uses the \Dataset{} name as the key.
To be effective, the graph-rewriting utility must implement mechanisms to
\begin{inlineenum}
  \item get a node's performance parameter (prefetching, parallelism),
  \item set a node's parallelism parameter, and
  \item insert a new node after the selected node (caching, prefetching)---each of
which is a standard graph operation.
\end{inlineenum}

\textbf{Automatic Optimizer Tool.}
On top of the same graph-rewriting code-base, we built a user-friendly pipeline optimizer tool, which
optimizes over parallelism, prefetching, and caching by rewriting the
\texttt{Graph}.
The pipeline optimizer is not only more extensive in its automatic optimizations, but
is meant to be faster than interactively finding bottlenecks.
For the pipeline-optimizer tool, there are three logical passes: LP-parallelism
optimization, prefetching insertion, and caching insertion.
Each of these logical passes can be executed independently or fused; by default
\plumber{} does 2 iterations of the passes, so that the estimated rates more
closely reflect the final pipeline's performance (recall that the empirical rates
shift slightly as parallelism is changed).

\textbf{Tracing Time.}
\plumber's tracing is able to return anytime, but longer tracing covers a larger
fraction of the (possibly resampled) dataset.
As slower pipelines generally require more tracing time to converge,
\plumber{} provides an option which stops when the difference between successive
throughput estimates drops below a small threshold---less than 5 minutes for
$1\%$ threshold in the pipelines we observe.

\begin{figure}
  \pythonexternal{Python/plumber_api.py}
  \vspace{-10pt}
  \caption{Python pseudo-code for optimizing a pipeline via annotations.
  \plumber{} will trace both caching paths and pick the one best suited for
  runtime conditions (e.g., memory allows caching).
  We note that the cached-path will typically have user-defined caching, though
  \plumber{} discards such performance-optimizations as \textit{suggestions}
  and inserts them itself, if possible, for each of the different
  code-paths---avoiding memory errors and duplicate caches.
  }%
  \label{fig:plumber_api}%
  \vspace{-10pt}%
\end{figure}

\textbf{Pick Best Queries.}
There are cases where two query algorithms implement the same
logical goal, with different performance characteristics.
For example, JPEG files can be decoded then randomly cropped, or the two can be fused.
The tradeoff is the former is amenable to caching (after decode), while the latter is faster
to decode---therefore, the optimal implementation depends on runtime conditions
(namely, memory capacity).
To get around \plumber's lack of a logical query system, \plumber{} allows users
to specify multiple pipelines, each with the same signature.
In practice, this is on the order of two pipelines (keeping the choice fast),
and \plumber{} will automatically trace both and apply any optimizations it can
before picking the fastest pipeline.
The optimization shown in Figure~\ref{fig:plumber_api}
is a difficult optimization to perform for an online tuner, like \autotune,
because, even with both pipelines to inspect, the effect of caching does not
kick in until a \textit{whole epoch into the training}.
\plumber{}, knowing that cache cold-start is a factor, can simulate the steady-state effects of caching
by truncating the cached data with advanced rewriting, yet still return the
normal dataset.
We use these annotations for our end-to-end ResNet experiments, which feature
the fused decode and crop example in the figure.

\subsection{Detecting Random UDFs for Caching}%
\label{subsec:random_udfs}%
For caching operators, \plumber{} must reason about both size and correctness
constraints.
In addition to size constraints (\S\ref{sec:system}), an operation may not be
cacheable if it is randomized; thus, \plumber{} performs additional analysis to
see if the transitive closure of the UDFs defined in an operation is random.
Specifically, we are interested in the relation: if a function, $f$, accesses a
random seed, $s$: $f \xrightarrow{} s$.
The transitive closure, $f \xrightarrow{+} s$, measures if \textit{any} child functions
of $f$ touch a random seed.
If $f \xrightarrow{+} s$ is true, then we cannot cache $f$ or any operations following
it.
This simple relation can be computed via a graph traversal and holds nearly
always,
as random seeds are necessary in implementations that enable determinism for
reproducibility.

\section{Additional Experimental Details.}
We provide additional experiments probing local behavior in this section, the cost
of profiling, and absolute speedup graphs of the end-to-end results provided in
the main text.

\subsection{End-to-End.}
The absolute end-to-end results are shown in Figure~\ref{fig:gcp_all_absolute}.
We include a MultiBoxSSD experiment, MultiBoxSSD (48), with only
half of the cores enabled for scheduling, demonstrating additional gains for
\plumber{} when less resources are available.
As mentioned in the main text, ResNetLinear is compared to conservatively
because \autotune{} sometimes allocates only 1 parallelism to reading I/O, which
results in only 6012 images/second rather than the 9230 images/second provided
by using the suggested pipeline parameters for file I/O.

\begin{figure}%
\centering%
\resizebox {0.99\linewidth} {!} {
\begingroup%
\makeatletter%
%
\makeatother%
\endgroup%
 }
\caption{
  Absolute throughputs in samples per second of the end-to-end model training experiments on
  TPUv3-8.
}%
\label{fig:gcp_all_absolute}%
\end{figure}

\begin{figure}
  \begin{subfigure}[t]{0.50\linewidth}
    \leftRightCrop{0.015}{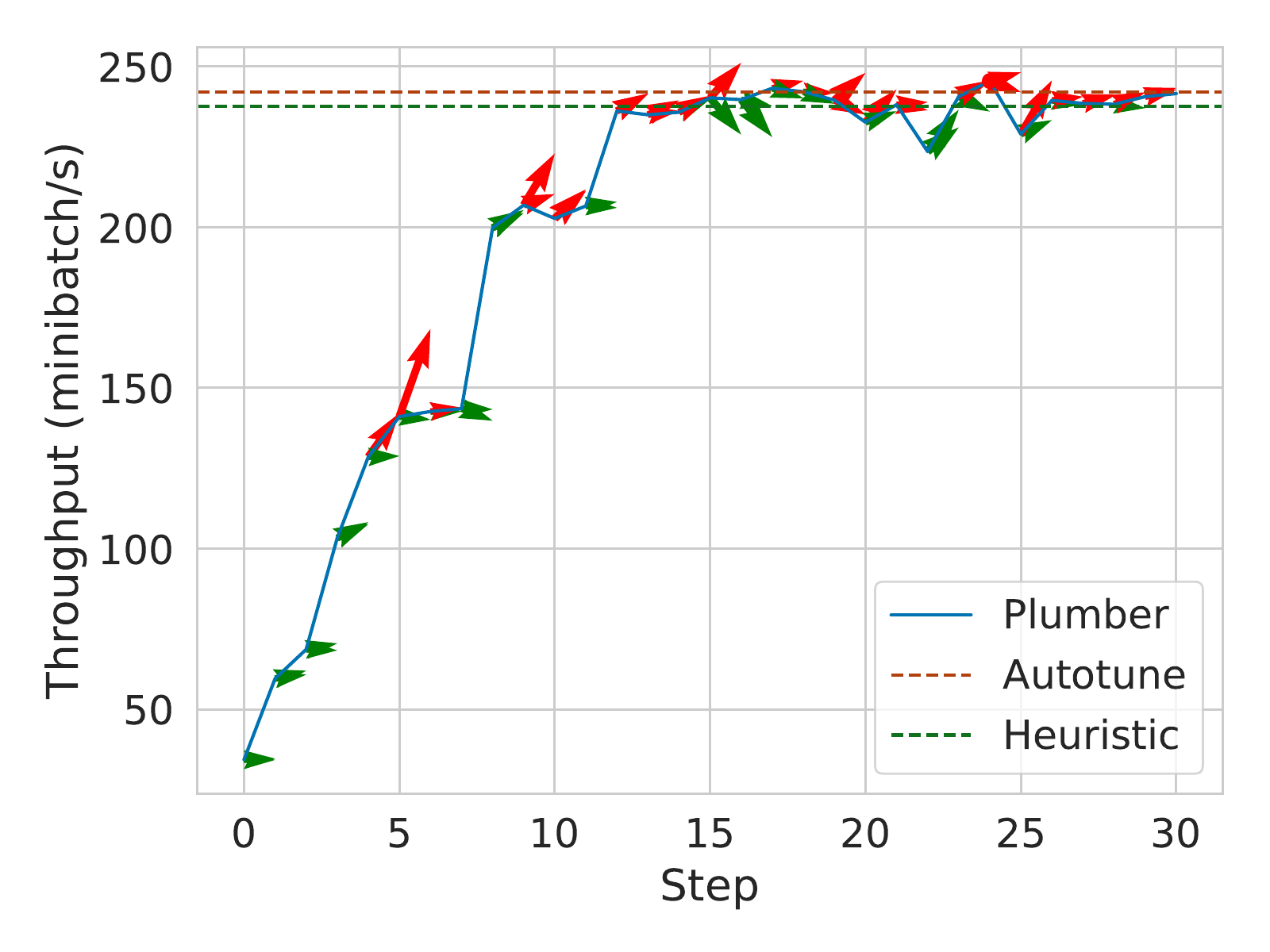}{1.06}{0.01}
    \caption{MultiBoxSSD (Setup A)}
  \end{subfigure}%
  \begin{subfigure}[t]{0.495\linewidth}
    \leftRightCrop{0.07}{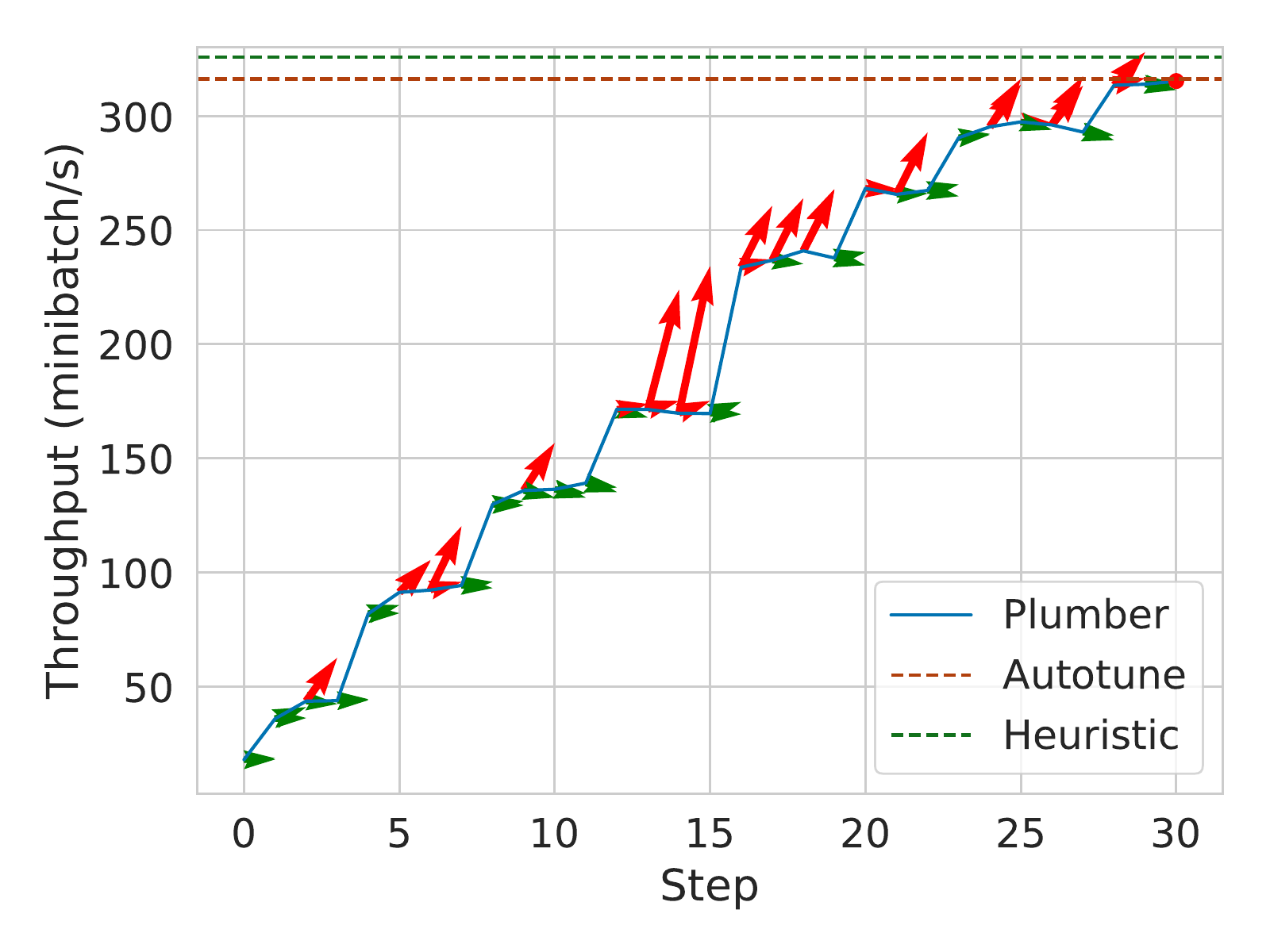}{1.06}{0.01}
    \caption{MultiBoxSSD (Setup B)}
  \end{subfigure}%
  \vspace{-5pt}
  \caption{
    MultiBoxSSD with local perturbations. \plumber{} is able to find optimal
    \Dataset{} choices when \Datasets{} are different with respect to
    performance. MultiBoxSSD exhibits two bottlenecks alternating every 4 steps, resulting in confusion at the steps.
    There are diminishing returns to pursuing an existing bottleneck, which slightly lag behind the rates measured.
  }%
  \vspace{-15pt}%
  \label{fig:ssd_convergence}%
\end{figure}

\subsection{Local Behavior.}
While \plumber{} may be able to converge to a good solution 2--3$\times$ faster than an uninformed user, it is worth knowing how \textit{optimal} are \plumber's predictions.
While we cannot enumerate all possible convergence paths, we can test if there was a better \Dataset{} selection to be made by \plumber{}.
To test this, we sample three one-step deviations from \plumber's recommended
action. We highlight MultiBoxSSD in Figure~\ref{fig:ssd_convergence}, which exhibits many transitions between bottlenecks.
For MultiBoxSSD, we observe similar bottleneck behavior as ResNet; \plumber{} prioritizes the
image processing operation and alternates to the \texttt{TFRecord}
parsing operation every 4 steps (nearly twice as often as ResNet cycles).
For ResNet, we observe local optimality except at the bottleneck transitions (once every 8 steps).
For the other datasets, we don't observe any significant mispredictions before convergence.
Across the datasets, we observe that both transition regions and resource
saturation cause measured rates to become \textit{correlated}---the rates begin
spiking/oscillating together, which makes ranking \Datasets{} unpredictable
On ResNet, for example, we observe oscillations begin at 90\% of peak performance.
Therefore, choosing the bottleneck node among similarly bottlenecked nodes is ambiguous.
For workloads with larger dynamic ranges between slower nodes, performance is nearly always optimal.

\subsection{Cost of Profiling}%
\label{sec:cost}
We evaluate the cost of running \plumber{} compared to an unmodified
\TensorFlow{}
2.4.1 build using the \heuristic{} configuration, which 1) stresses the
system and 2) avoids \autotune-related overheads.
On setup A, the average slowdown across the 5 pipelines is 5\% and is driven
entirely by Transformer/GNMT, which have a slowdown of 19\%/21\%, respectively.
Thus, the tracing overhead grows with less work per element (motivating a
batched execution engine).
On setup B, the story is similar, though the effect is more pronounced (likely
due to increased overhead from timer system calls);
average slowdown is 10\% and is 17\%/36\% across
Transformer/GNMT, respectively.
The overhead is still
lower than the \TensorFlow{} Profiler, which incurs a 7--15\% slowdown on vision workloads
and 40\% slowdown on the text workloads on system A.

\section{Dataset Details.}
We summarize the MLPerf datasets below.
\textbf{ResNet50}~\cite{he2016deep} is an image
classification task selecting between $1000$ categories of the ImageNet~\cite{deng2009imagenet} dataset. %
\textbf{Mask RCNN}~\cite{ren2016faster} is an object detection and segmentation task over the MSCOCO~\cite{lin2014microsoft} dataset.
\textbf{MultiBoxSSD}~\cite{liu2016ssd} is a real-time object detection and segmentation task over the MSCOCO dataset.
\textbf{Transformer}~\cite{vaswani2017attention} is a neural machine
translation task, which converts a sequence of words from a source to
a target language using the WMT English-to-German dataset~\cite{WMT17}.
\textbf{GNMT}~\cite{wu2016google} is a task similar to Transformer using the WMT 2016~\cite{WMT16} dataset and a different pipeline.
The ImageNet dataset is 148GB, COCO is 20GB for both MaskRCNN and MultiBoxSSD, the processed WMT data for
Transformer is 1.2 GB, and the processed WMT data for GNMT is 1.9GB.

\section{Artifact Appendix}%
\label{sec:artifact}%

\subsection{Abstract}
We provide three logical artifacts: a \TensorFlow{} fork, an application-layer
\texttt{Python} library on top of \TensorFlow, and example pipelines used for evaluation.
These artifacts allow a user to use \plumber{} to diagnose their own pipelines as
well as recreate the main results of the paper.
To validate the functionality of the \TensorFlow{} and \texttt{Python} library, we provide a
simple notebook that requires minimal resources and can be run quickly.
To validate microbenchmark functionality, a machine with 16GiB RAM and a 4+ core
CPU is sufficient, while end-to-end validation requires access to (proprietary) Google Cloud
TPUs.
Microbenchmarks and end-to-end runs require running a data pipeline with
optional model attached for a number of iterations, and the main metric is the
rate at which those iterations are processed (discussed in paper).
The primary skills necessary are experience with \TensorFlow{} and \Jax{} and familiarity with
\tfdata{} in addition to TPUs.

\subsection{Artifact Check-List (Meta-Information)}

{\small
\begin{itemize}
  \item {\bf Program: Python, C++}
  \item {\bf Compilation: C++, Bazel}
  \item {\bf Data set: ImageNet, COCO, WMT16, WMT17}
  \item {\bf Hardware: CPU-only server, TPUv3--8 VM}
  \item {\bf Metrics: Throughput}
  \item {\bf Experiments: Machine Learning Training, Microbenchmarks}
  \item {\bf How much disk space required (approximately)?: 250GB }
  \item {\bf How much time is needed to prepare workflow (approximately)?: A few
    hours to prepare the datasets (e.g., TFRecord conversion) and a few hours to
    compile \TensorFlow.}
  \item {\bf How much time is needed to complete experiments (approximately)?:
    The microbenchmarks complete in a few hours each, with ResNet/ImageNet being
    the slowest, due to the number of optimization steps it takes. The end-to-end results
    can finish in a few hours each, except for RCNN, which can take about a day
    due to long epochs.}
  \item {\bf Publicly available?: Yes}
  \item {\bf Code licenses (if publicly available)?: Apache License 2.0}
\end{itemize}
}

\subsection{Description}

\subsubsection{How Delivered}
We provide two open-source GitHub repositories.
The \TensorFlow{} fork is available at:
\url{https://github.com/mkuchnik/PlumberTensorflow}.
The \plumber{} \texttt{Python} application-level library and experiment files are available at:
\url{https://github.com/mkuchnik/PlumberApp}.
The latter repository contains the instructions for using both the \plumber{}
\TensorFlow{} fork and the \texttt{Python} application-level library.

\subsubsection{Hardware Dependencies}
Microbenchmarks should run on any hardware that runs \TensorFlow. However,
end-to-end results require the use of Google Cloud TPU VMs (TPUv3--8).

\subsubsection{Software Dependencies}
We use \Jax, \TensorFlow, \texttt{Python}, and a proprietary but publicly
available library for compiling against TPUs.
A non-exhaustive list of open-source \texttt{Python} packages is:
\texttt{numpy},
\texttt{matplotlib},
\texttt{seaborn},
\texttt{cvxpy}, and 
\texttt{graphsurgeon}.

\subsubsection{Data Sets}
We use ImageNet~\cite{deng2009imagenet}, COCO~\cite{lin2014microsoft}, WMT16
(EN-DE)~\cite{WMT16}, and WMT17 (EN-DE)~\cite{WMT17}.

\subsection{Installation}
Build and install \TensorFlow{} using the publicly available
documentation.
The \texttt{Python} library's source code is packaged to be installed as a \texttt{Python}
wheel.

\subsection{Experiment Workflow}
For microbenchmarks, run an iterative graph-rewriting script that compares random
optimizations to \plumber's recommendations.
The throughput of the pipeline is measured (e.g., minibatches per second) as
well as the estimated rate according to \plumber{} and \autotune{}.
These curves are then compared to naive and strong baseline methods
(\heuristic{} and \autotune{}) of tuning the
pipeline.
For end-to-end experiments, run a subset of ML training using naive, baseline
methods, and \plumber's optimized pipeline.
The throughput of the end-to-end training is measured (e.g., minibatches per second).

\subsection{Evaluation and Expected Result}
For microbenchmarks, the throughput curve of random optimizations should be
worse than that of \plumber's recommendations.
For throughput, \plumber{} should beat naive policies and match strong
baselines.
For estimated rates, \plumber's estimated rates should be bounded, unlike
\autotune{}'s.
For end-to-end results, \plumber{} should match or exceed strong baselines on
most pipelines.

\subsection{Methodology}

Submission, reviewing and badging methodology:

\begin{itemize}
  \item \url{http://cTuning.org/ae/submission-20190109.html}
  \item \url{http://cTuning.org/ae/reviewing-20190109.html}
  \item \url{https://www.acm.org/publications/policies/artifact-review-badging}
\end{itemize}

\end{document}